%% file: 0_main.tex
\documentclass[10pt,journal,compsoc]{IEEEtran}
\ifCLASSOPTIONcompsoc
  \usepackage[nocompress]{cite}
\else
  \usepackage{cite}
\fi

\ifCLASSINFOpdf
  \usepackage[pdftex]{graphicx}
  \graphicspath{{../pdf/}{../jpeg/}}
  \DeclareGraphicsExtensions{.pdf,.jpeg,.png}
\else
  \usepackage[dvips]{graphicx}
  \graphicspath{{../eps/}}
  \DeclareGraphicsExtensions{.eps}
\fi

\usepackage{amsthm}
\theoremstyle{definition}
\newtheorem{definition}{Definition}[section]
\newtheorem{example}{Example}[section]
\usepackage{framed}
\usepackage{multirow}
\usepackage{algorithm}
\usepackage{algorithmicx}
\usepackage{algpseudocode}
\usepackage{xcolor}
\usepackage{subfigure}
\usepackage{footnote}
\newcommand{\methode}{ADF}
\newcommand{\method}{ADF }

\newcommand{\datasets}{5 }
\newcommand{\benchmarks}{22 }
\newcommand{\tabulardata}{3 }
\newcommand{\textdata}{2 }
\newcommand{\questions}{4 }
\newcommand{\idi}{discriminatory sample }
\newcommand{\idie}{discriminatory sample}
\newcommand{\aeq}{AEQUITAS }
\newcommand{\aeqe}{AEQUITAS}
\newcommand{\tm}{THEMIS }
\newcommand{\tme}{THEMIS}

\usepackage{amsfonts}
\newcommand{\yes}{\checkmark}
\usepackage{pifont}
\newcommand{\no}{\hspace{1pt}\ding{55}}

\begin{document}
\title{Automatic Fairness Testing of Neural Classifiers through Adversarial Sampling}

\author{Peixin Zhang,
        Jingyi Wang,
        Jun Sun, 
        Xinyu Wang,
        Guoliang Dong,\\
        Xingen Wang,
        Ting Dai
        and Jin Song Dong
\IEEEcompsocitemizethanks{\IEEEcompsocthanksitem Peixin Zhang is with Zhejiang University, P.R. China. E-mail: pxzhang94@zju.edu.cn
\IEEEcompsocthanksitem The corresponding authors Jingyi Wang, Xinyu Wang are with Zhejiang University, P.R. China. E-mail: {wangjyee,wangxinyu}@zju.edu.cn
\IEEEcompsocthanksitem Jun Sun is with Singapore Management University, Singapore. E-mail: junsun@smu.edu.sg
\IEEEcompsocthanksitem Guoliang Dong, Xingen Wang are with Zhejiang University, P.R. China. E-mail: {dgl-prc,newroot}@zju.edu.cn
\IEEEcompsocthanksitem Ting Dai is with Huawei International Pte. Ltd., P.R.China. E-mail: daiting2@huawei.com
\IEEEcompsocthanksitem Jin Song Dong is with National University of Singapore, Singapore. E-mail: dcsdjs@nus.edu.sg}}


\IEEEtitleabstractindextext{%
\begin{abstract}
Although deep learning has demonstrated astonishing performance in many applications, there are still concerns about its dependability. One desirable property of deep learning applications with societal impact is fairness (i.e., non-discrimination). Unfortunately, discrimination might be intrinsically embedded into the models due to the discrimination in the training data.  As a countermeasure, fairness testing systemically identifies discriminatory samples, which can be used to retrain the model and improve the model's fairness. Existing fairness testing approaches however have two major limitations. Firstly, they only work well on traditional machine learning models and have poor performance (e.g., effectiveness and efficiency) on deep learning models. Secondly, they only work on simple structured (e.g., tabular) data and are not applicable for domains such as text. In this work, we bridge the gap by proposing a scalable and effective approach for systematically searching for discriminatory samples while extending existing fairness testing approaches to address a more challenging domain, i.e., text classification. Compared with state-of-the-art methods, our approach only employs lightweight procedures like gradient computation and clustering, which is significantly more scalable and effective. Experimental results show that on average, our approach explores the search space much more effectively (9.62 and 2.38 times more than the state-of-the-art methods respectively on tabular and text datasets) and generates much more \idie s (24.95 and 2.68 times) within a same reasonable time. Moreover, the retrained models reduce discrimination by 57.2\% and 60.2\% respectively on average.
\end{abstract}

\begin{IEEEkeywords}
Deep learning, fairness testing, individual discrimination, gradient
\end{IEEEkeywords}
}

\maketitle

\IEEEdisplaynontitleabstractindextext

\IEEEpeerreviewmaketitle

\input{1_Introduction}
\input{2_Preliminary}
\input{3_Overview}
\input{3_Methodology}
\input{4_Experiment}
\input{5_Discussion}

\input{6_Related_Work}

\input{7_Conclusion}

\section*{Acknowledgments}
This research was supported by the Key-Area Research and Development Program of Guangdong Province (Grant No.2020B0101100005). This research was also supported by the Key Research and Development Program of Zhejiang Province (Grant No.2021C01014), the Fundamental Research Funds for the Zhejiang University NGICS Platform, the Guangdong Science and Technology Department under Grant No. 2018B010107004, and Ministry of Education, Singapore (Project MOET32020-0004, T2EP20120-0019, and T1-251RES1901).

\bibliographystyle{IEEEtran}
\bibliography{Fairness}
\vfill
\end{document}

%% file: 1_Introduction.tex
\IEEEraisesectionheading{\section{Introduction}\label{sec:introduction}}

\IEEEPARstart{I}{n} recent years, deep learning (DL) has been applied to various areas of our daily life, e.g., face recognition~\cite{face_recognition}, fraud detection~\cite{fraud_detection}, and natural language processing (NLP)~\cite{toxic}. As DL is increasingly connected with our society, we cannot simply regard it as a mathematical abstraction, but rather as a society-technical system~\cite{fairness_ml}. In other words, besides testing the accuracy (i.e., a quantification for evaluating the effectiveness of mathematical approximation) of the DL model, we also need to take ethical properties into consideration. Among them, fairness (i.e., non-discrimination) is one of the properties that raise the most heightened public concern~\cite{trust_ai}, especially in minority and vulnerable groups. In ~\cite{fairtest}, Tramer \emph{et al.} demonstrated that the existing discrimination in our society may be present in the training data and learned by DL models unintentionally. Worse yet, discrimination in DL is often more `hidden' than that of traditional decision-making software since it is still an open problem on how to interpret DL models. Therefore, it is crucial to have systematic methods for automatically identifying discrimination in DL systems.

Various forms of discrimination exist in the machine learning literature, including group discrimination~\cite{group_discrimination,verification} and individual discrimination~\cite{fairness,counterfactual}. Discrimination is often defined over a set of protected attributes\footnote{We use `protected'/`sensitive' and `attribute'/`feature' interchangeably.}, such as age, race and gender. Intuitively, discrimination happens when a machine learning model makes different decisions for different \emph{individuals} (individual discrimination) or \emph{subgroups} (group discrimination) differentiated only by protected attributes. Note that the set of protected attributes is often application-dependent and given in advance. We refer the readers to~\cite{science} for detailed definitions of fairness.

\begin{figure}[t]
\centering
\includegraphics[scale=0.5]{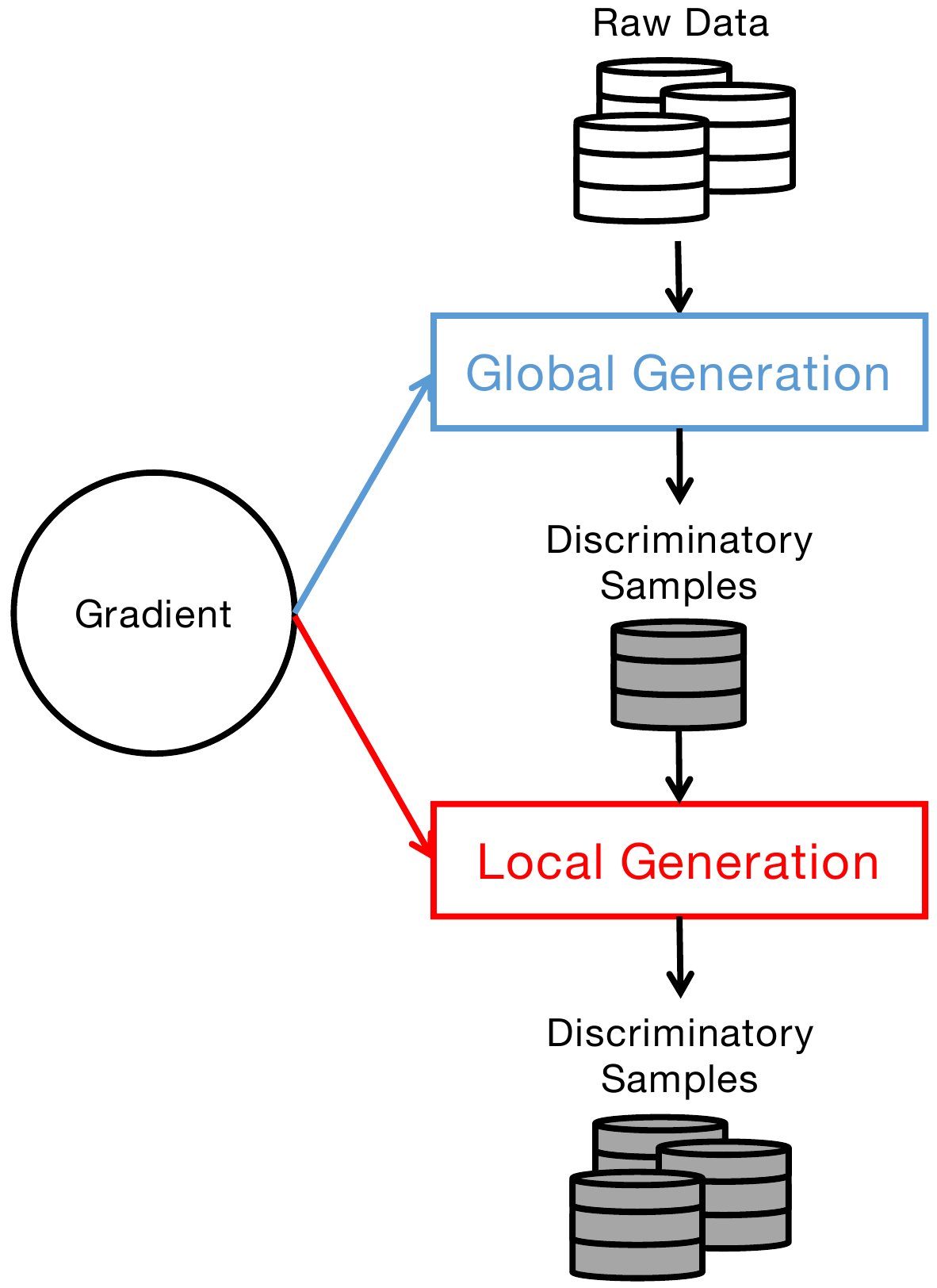}
\caption{An overview of \method.}
\label{fig:overview}
\end{figure}

In this work, we focus on the problem of individual discrimination and aim to develop a systematic and scalable approach for identifying/generating \idie s for DL models, i.e., a pair of samples that differ only by some protected features but have different labels. Note that in a given DL model, it is often inadequate to identify a few samples demonstrating the existence of individual discrimination. Instead, we need to generate as many discrimitive samples as possible and use them to improve the fairness (i.e., mitigate the discrimination) of the DL model by retraining. In the software engineering literature, there have been multiple attempts on the problem~\cite{themis,aequitas,sg}. For instance, \tm~\cite{themis} randomly samples each attribute within its domain and identifies those discriminative samples. \aeq~\cite{aequitas} improves the testing effectiveness with a two-phase (global and local) search. SG~\cite{sg} combines the local explanation model~\cite{lime} and symbolic execution~\cite{concolic_testing} to increase the diversity and the number of discriminatory samples.

Existing approaches are developed mostly for traditional machine learning models, i.e., logistic regression, support vector machine, and decision tree. Although they could be applied to DL models, experiment results show that their performance on DL models is far worse than that on traditional models, and are far from being practical. We discuss the detailed shortcomings for each approach in Section~\ref{qualitative}. Besides, the aforementioned algorithms only work on simple tabular data and are not applicable to complex data formats (e.g., text for NLP tasks). The difficulty is two-fold. Firstly, the features of most tabular data are well-designed and relatively independent from each other, which allows approaches such as \tm and \aeq to generate new values independently for each feature. For text, each token (i.e., word or punctuation) is tightly embedded in its context. As a result, the \idie s generated by \tm and \aeq (if they manage to work) would unlikely be valid. Secondly, the text for NLP models is often pre-processed through an embedding procedure, which converts tokens to distributed and non-continuous vectors. Such vector inputs further make constraint solving based approaches such as SG inapplicable since it is challenging to solve constraints on such vectors.

To address the challenges, we propose a novel scalable gradient-based algorithm named \textit{A}dversarial \textit{D}iscrimination \textit{F}inder (\methode) for generating discrimitive samples, which is specifically designed for DL models. The gradient is an effective tool to craft test inputs for DL models. It can be calculated efficiently and offers intuitive guidance on how a model's outcome changes with respect to certain attributes. It is proved to be useful in various DL-related tasks. It is noticeably utilized in recent works to generate adversarial samples~\cite{fgsm,bim,asc,jsma,deepxplore}, i.e., samples which are only slightly different from a given one in the training set yet result in very different model prediction. Inspired by these works, we use the gradient as an effective way for searching discrimitive samples.

An overview of \method is presented in Figure ~\ref{fig:overview}. \method has two phases, i.e., global generation and local generation. The goal of the global generation is to increase diversity in the generated \idie s. During the global stage, \method selects samples in the raw dataset and uses the gradients to guide the search of \idie s, by maximizing the difference between the model predictions of two similar samples. Then local generation takes these identified \idie s as seeds to acquire more discriminatory samples by exploring their neighbors. Here, \method utilizes the gradient in a different way to identify \idie s which are minimally different from the seeds while maintaining their models' outputs. Note that \method is generative, i.e., it may generate samples that are not in the original dataset. In order to make sure the generated samples are valid, we utilize a clip function on tabular datasets and semantically similar words on text datasets in both global and local generation phases.

\method has been implemented as in a self-contained toolkit. Our experiments on \benchmarks real-word benchmarks (including both tabular and text datasets) show that \method signficantly outperforms the baseline methods. For tabular data, \method explores 8 times more input space and generates 24 times more \idie s on average than \aeqe. Comparing with SG, \method has an average of 324\% and 32\% higher success rate in global generation and local generation. Furthermore, \method is around 2 and 7 times more efficient than \aeq and SG. For text data, \method generates 2.68 times \idie s and has a 9.01\% more success rate than random perturbation on average within a same reasonable time. 

In summary, we make the following contributions.
\begin{itemize}
	\item We propose a systematic and scalable algorithm \method for generating \idie s of DL models efficiently and effectively based on gradient.
	\item We address the challenges of applying fairness testing for complex data format (beyond tabular data), i.e., text.
	\item We evaluate \method with \benchmarks benchmarks on \datasets datasets (including \tabulardata tabular datasets and \textdata text datasets). Our experiment shows \method is significantly more effective and efficient in generating \idie s than state-of-the-art methods, e.g., AEQUITAS~\cite{aequitas} and Symbolic Generation~\cite{sg}.
	\item We implement and publish \method as an online end-to-end service~\cite{appendix} to improve the fairness of DL models.
\end{itemize}

This paper is an extension of our previous publication~\cite{white_fairness} by applying the original idea of \method to a much more challenging domain, i.e., text. In particular, we present the difficulties of why existing fairness testing approaches are infeasible for text and our approaches to address the challenges. We also provide several additional running examples and much more extensive experiments on text datasets to demonstrate the effectiveness of our approaches.

We organize the rest of the paper as follows. We first provide the background of DL and gradient-based adversarial attacks in Section~\ref{sec:preliminary}. We present the details of the \method in Section~\ref{sec:methodology}. In Section~\ref{sec:experiment}, we evaluate our approach on multiple real-world datasets and demonstrate its effectiveness and efficiency. We review the related works in Section~\ref{sec:relatedwork}. Finally, in Section ~\ref{sec:conclusion}, we conclude our work and discuss the potential future directions.

%% file: 2_Preliminary.tex
\section{Preliminary}
\label{sec:preliminary}
In this section, we review relevant background.

\subsection{Deep Learning}
\label{subsec:dl}
We target two kinds of commonly used DL models, i.e., Feedforward Neural Network (FNN) and Recurrent Neural Network (RNN).

\noindent\textbf{FNN} A FNN contains an input layer, an output layer, and multiple hidden layers. We denote these layers as $L = \{ L_{j}|j \in \{0, \dots, J \} \}$, and the $j$-th layer are consisted of $|L_{j}|$ neurons. A FNN computes the outcome of each neuron by applying activation function (e.g., Sigmoid, tanh, or relu~\cite{relu}) $\phi$ to the weighted sum of the outputs of all the neurons in its previous layer.

\noindent\textbf{RNN} RNN is a variant of the traditional neural network, e.g., FNN, which is designed for modeling sequential data. An RNN takes a sequence $x=x_0 x_1\cdots x_n$ as input and produces a sequence $o=o_0 o_1 \cdots o_m$ as output. We use $x[i]$ to denote the $i$-th element of $x$ and $|x|$ to denote the length of $x$. RNN `memorizes' what has been calculated so far through a set of hidden states $H$. At each time step $t$, the current hidden state $h_t$ and the output $o_t$ are calculated as follows.
\begin{itemize}
	\item $x_t$ is the input at time $t$.
	\item $h_t$ is the hidden state (which is also known as the memory unit) at time $t$. It can be calculated based on the last hidden state and the current input: $h_t = \phi(U\cdot x_t+W\cdot h_{t-1})$, where $\phi$ is the activation function, $U$ and $W$ are the weights of current input and previous state respectively. Specifically, $h_{-1}$ is initialized to be 0 for calculating the first hidden state.
	\item $o_t$ is the output at time $t$ which is obtained by $o_t = \psi(V\cdot h_t)$, where $\psi$ is normally a Softmax function, $V$ is the weight of the current hidden state.
\end{itemize}

In this work, we focus on DL classifier $\mathcal{D}: X \to Y$, i.e., for a given sample $x \in X$, a DL model outputs a predicted label $y \in Y$ which has the highest probability. Note that we use $\theta$ to denote the set of its parameters. 

\subsection{DL Ethics}
\label{subsec:ethics}

Given that DL models are increasingly applied in applications with significant societal impact, they must act following the formal (i.e., Law) and informal norms (i.e., Ethics) held for human beings. In other words, in addition to improving the accuracy of the models, we hope to develop Trustworthy AI, since only when human beings trust AI, can they harvest the benefits of this technology confidently and adequately.

The High-level Expert Group on Artificial Intelligence (AI HLEG) of the European Commission lists $10$ fundamental requirements of Trustworthy AI~\cite{trust_ai}, e.g., \textit{robustness} requires DL models to handle errors or inconsistencies in its life cycle~\cite{fgsm,jsma,asc,hotflip,iceccs,model_mutation}; \textit{transparency} aims to help developers and users better understand the logic and reasons in the decision-making process~\cite{lime,lemna}; and \textit{privacy} ensures that all personal information used and generated in the intersection between users and DL models are not leaked~\cite{data_privacy1,data_privacy2}. In addition, \textit{fairness (non-discrimination)} is a new research hotspot in the field of AI ethical principles in recent years, which attempts to avoid DL models that intentionally or unintentionally marginalize certain minorities.

So far, there is no commonly agreed definition of fairness, despite that many definitions have been proposed in the literature, e.g., Demographic Parity and Equalised Odds. In this work, we focus on individual fairness, i.e., given two samples that only differ by the protected attributes, a model must output the same label. The formal definition of individual fairness is in the next section.

\subsection{Individual Discrimination}
\label{subsec:id}
We denote $X$ as a dataset and its set of elements by $E = \{E_{1}, E_{2}, \dots, E_{n}\}$, i.e., an element is an attribute for tabular data and a token for text data. Assuming each element $E_i$ has a valuation domain $\mathbb{I}_i$, the input domain is then $\mathbb{I} = \mathbb{I}_{1} \times \mathbb{I}_{2} \times \dots \times \mathbb{I}_{n}$, which denotes all the possible combinations of element valuations. Note that given an unknown sample, $n$ is a constant and each attribute has its own domain for tabular data, whereas for text data, $n$ varies with inputs and the input space is the entire corpus. Further, we use $P \subset A$ to denote a set of protected attributes like race and gender and $NP$ to denote the set of non-protected attributes. It is easy to obtain the protected attributes and their value domains for tabular data. However, for the text, we need to identify a list of features that are often associated with fairness issues as protected attributes and define the value domain for each one manually. For instance, one of the protected attributes is `country' and it has 160 possible values. Note that this is a one-time effort that is manageable and reusable. The details of the identified protected attributes and their value domains are available at~\cite{appendix}. A DL model $\mathcal{D}$ trained on $X$ may contain individual discrimination which is defined as follows.

\begin{definition} 
\label{de:defidi}
Let $x=\{x_{1}, x_{2}, \dots, x_{n}\}$, where $x_{i}$ is the value of attribute $A_{i}$  be an arbitrary sample in $\mathbb{I}$. We say that $x$ is a \idi of a model $\mathcal{D}$ if there exists another data sample $x'\in\mathbb{I}$ which satisfies the following conditions:
\begin{itemize}
      \item $\exists p \in P, s.t., x_{p} \neq x_p'$;
      \item $\forall q \in NP, x_{q} = x'_{q}$;
      \item $\mathcal{D}(x) \neq \mathcal{D}(x')$
\end{itemize}
Further, $(x,x')$ is called a \idi pair.
\end{definition}

\subsection{Gradient-based Adversarial Attack}
\label{subsec:gba}
In~\cite{intriguing}, Szegedy \emph{et al.} demonstrated that DL models are vulnerable to adversarial samples, which are crafted by introducing perturbations on the original normal samples to mislead the decisions of the model. Various kinds of adversarial attacks have been proposed in the past few years~\cite{fgsm,bim,jsma,asc,textbugger,deepxplore}, and gradient-based methods are one of the fastest and most straightforward ways. The gradient of the prediction $y$ with respect to the input $x$ is defined as:  
\begin{equation}
\label{eq:gradient}
G(x,y) = \frac{\partial J(\theta, x, y)}{\partial x}
\end{equation}
where $J$ could be any objective function related with the output, e.g., the loss function~\cite{fgsm,bim,jsma,robot}, the logits (i.e., the input values of the last softmax layer)~\cite{logits_gradient,asc}. In the following, we briefly introduce two representative gradient-based adversarial attacks for tabular data and text data.

\vspace{1mm}
\noindent\textbf{FGSM} In~\cite{fgsm}, Goodfellow \emph{et al.} proposed Fast Gradient Sign Method (FGSM) which perturbs the original input in the direction of the sign of the gradient of the model’s loss function with respect to the input attributes according to the following Equation:
\begin{equation}
\label{equation:fgsm}
x^{adv} = x + \epsilon \cdot \textbf{sign}(G(x,y)),
\end{equation}
where $\epsilon$ is a hyper-parameter to determine the perturbation degree.

\vspace{1mm}
\noindent\textbf{ASC} In~\cite{asc}, Papernot \emph{et al.} proposed Adversarial Sequence Craft (ASC) which iteratively replaces an attribute with a substitute value that impacts the model's outcome significantly until success or timeout. In other words, it chooses the value with the closest direction (i.e., the sign of the difference between it and the original value) to the one indicated by the gradient.

\vspace{1mm}
\noindent\textbf{HotFlip} In~\cite{hotflip}, Ebrahimi \emph{et al.} proposed a white-box adversarial attack method named HotFlip, which estimates the impact of each flip operation on one-hot input by the gradient, and utilizes the beam search to find the best perturbation with the highest loss.

\subsection{Problem Definition}
\label{subsec:pd} 
A model which suffers from individual discrimination may produce prejudiced decisions when a \idi is presented as input. Our problem is thus defined as follows. Given a dataset $X$, a set of protected attributes $P$ and a DL model $\mathcal{D}$, how can we effectively and efficiently generate \idie s for $\mathcal{D}$ so that we can retrain a DL model based on $X$ and the generated \idie s for more fair models?
This problem is challenging because we focus on complex DL models which renders existing methods ineffective.

%% file: 3_Overview.tex
\section{Overview}
\label{sec:me}
\begin{figure}[t]
\subfigure[Global Generation]{
\includegraphics[width=0.23\textwidth]{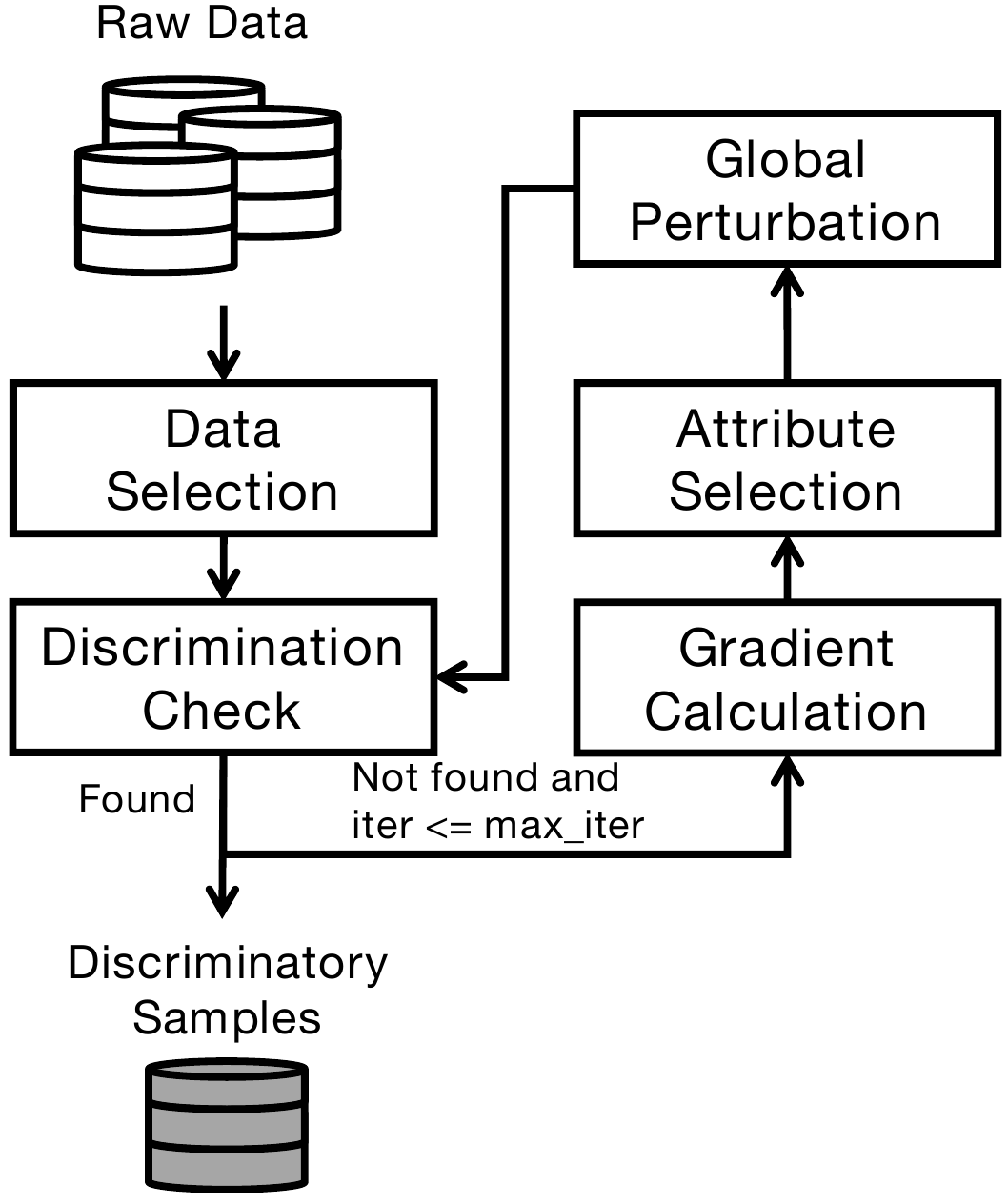}
\label{fig:ogg}
}
\subfigure[Local Generation]{
\includegraphics[width=0.23\textwidth]{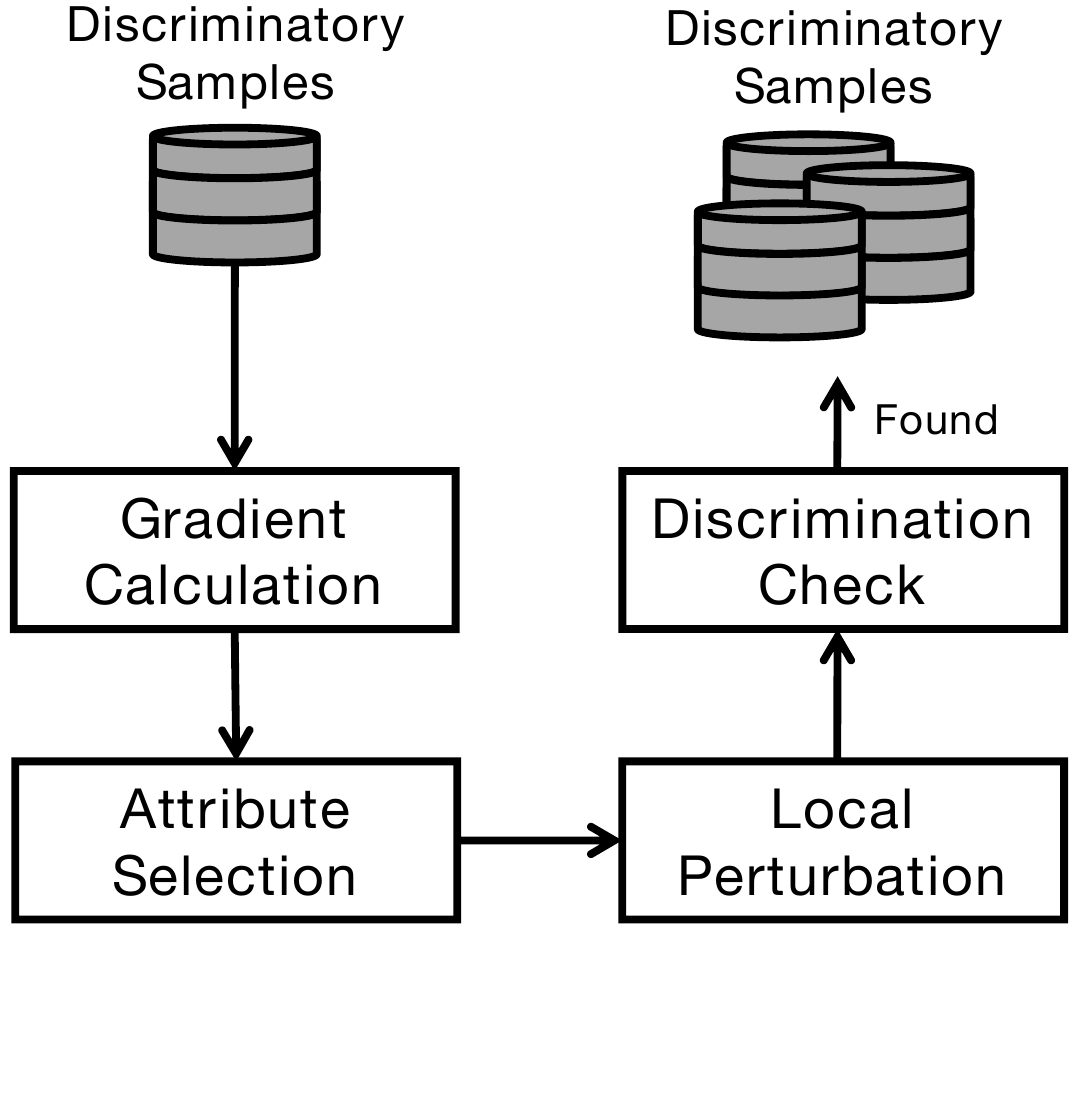}
\label{fig:olg}
}
\caption{The pipeline of \method.}
\label{fig:o}
\end{figure}

In this section, we present the pipeline of global and local generation of our method \method in Figure~\ref{fig:ogg} and ~\ref{fig:olg} respectively and utilize a toy example to explain the relevant notations.

\begin{figure}[t]
\centering
\includegraphics[scale=0.6]{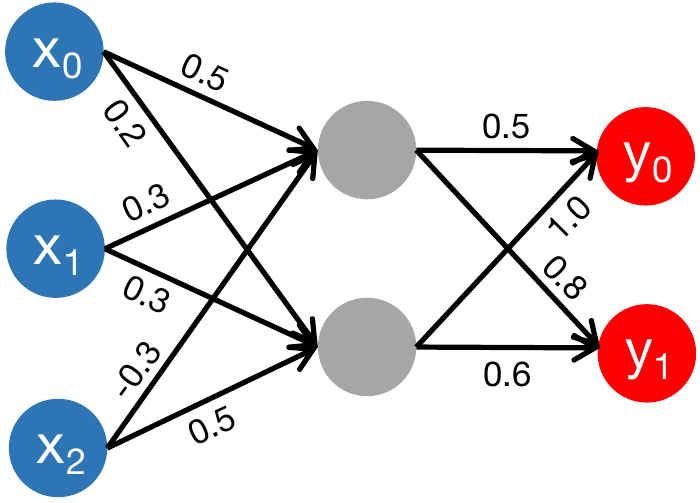}
\caption{Toy example of FNN without activation function.}
\label{fig:toy}
\end{figure}

Figure~\ref{fig:toy} depicts a one-layer FNN without activation function for a sensitive classification task. The input vector is $[x_0, x_1, x_2]$ (shown in blue), where all of them have continuous integer values and $x_2$ is the only protected attribute with two possible values, 0 and 1. $y_0$ and $y_1$ (shown in red) denote the output nodes. Besides, each value in the figure represents the corresponding weight between two linked neurons.

In the global phase, we first select seeds to search as many different parts of the input space as possible, e.g., $[0,1,0]$. Next, we follow Definition~\ref{de:defidi} to check if the output ($y_0$) of the seed is consistent when we only flip the sensitive attribute $x_2$. Since gradient indicates the direction and magnitude of the output change with respect to the perturbation on the input~\cite{fgsm}, we use gradient calculated according to the chain rule, e.g., $\partial y_0 / \partial x_0 = 0.5*0.5+1.0*0.2=0.45$, to search for \idie s more efficiently. Then we select the non-protected attribute(s) and perturb them by gradient, e.g., we choose the attribute $x_0$ (with larger gradient) and apply $x_0 = x_0 + 1 * sign(0.45)$, to acquire a sample that is more likely discriminatory. The new sample $[1,1,0]$ is used to start the next iteration. The global generation stops if a \idi is successfully generated or it times out.

In the local phase, we further search the neighborhood of identified \idie s to increase the quantity. Here, the gradient is computed to guide the selection of non-protected attribute(s) and perturbation to ensure that the search space is as close to the existing discriminatory sample as possible. Last, we check whether the new sample is discriminatory according to Definition~\ref{de:defidi}.

%% file: 3_Methodology.tex
\section{Methodology}
\label{sec:methodology}
In this section, we first present details of our approach \method and then a qualitative comparison between our approach and state-of-the-art fairness testing approaches.

\method generates \idie s in two phases, i.e., a global generation phase and a local generation phase. In the global generation phase, we aim to identify those \idie s near the decision boundary from the original dataset $X$, which then serve as the seed data for the local generation phase. In the local generation phase, we follow the intuition that samples in the neighborhood of those seed data are more likely to be \idie s to obtain more of them. Note that this intuition is inspired by recent research on the robustness of DL models~\cite{intriguing,input_mutation}. In the following, we introduce the two phases in detail with two running examples (from tabular and text dataset respectively).

\begin{example}
\label{exp:1.1}
We use the Census Income dataset\footnote{https://archive.ics.uci.edu/ml/datasets/adult} as a running example on tabular data to illustrate each step of our approach. The Census Income dataset is published in 1996, which is a commonly used dataset in the literature of fairness research~\cite{themis,aequitas,sg,google1,google2}. The task is to predict whether the income of an adult is above 50,000\$ based on their personal information. The dataset contains 32,561 training samples with 13 attributes each. The following shows a sample $x$.
$$x: [4, 0, 6, 6, 0, 1, 2, 1, \textcolor{red}{1}, 0, 0, 40, 100]$$
Note that all the attributes are category attributes (obtained through binning). Among the 13 attributes, there are multiple potential protected attributes, i.e., age, race, and gender. In the following, we assume the protected attribute is gender for simplicity, whose index in the feature vector is 8 (which is highlighted in red above). There are only two different values for this attribute, i.e., 0 representing female and 1 representing male. Given a model trained on the dataset, if changing 1 to 0 changes the prediction outcome by the model, we say that $x$ is a \idi for the model.
\end{example}

\begin{figure}[t]
\centering
\subfigure[Original sample]{
\includegraphics[width=0.35\textwidth]{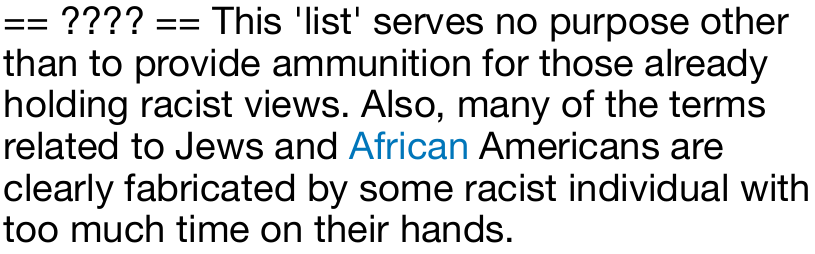}
\label{fig:os}
}
\subfigure[IDS identified by perturbation on non-IDS]{
\includegraphics[width=0.35\textwidth]{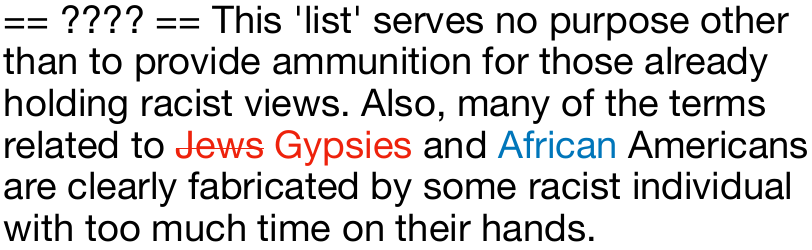}
\label{fig:gs}
}
\subfigure[IDS identified by perturbation on IDS]{
\includegraphics[width=0.35\textwidth]{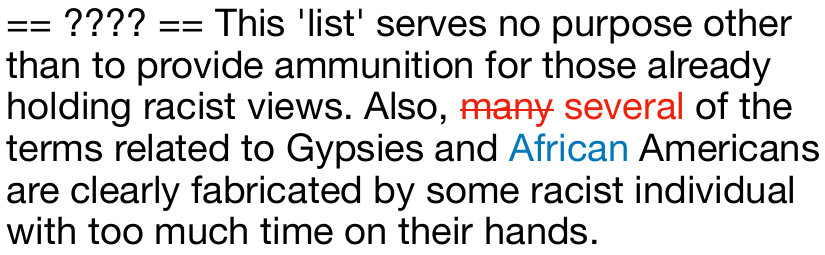}
\label{fig:ls}
}
\caption{Sample searching results}
\label{fig:ma2}
\end{figure}

\begin{example}
\label{exp:2.1}
We utilize the text dataset Wikipedia Comments\footnote{https://github.com/conversationai/unintended-ml-bias-analysis} as a running example on text data. This dataset is collected from Wikipedia Talk Pages for classification~\cite{wiki}. The classification task is to label each comment as either toxic or non-toxic. It was also adopted for measuring discrimination~\cite{abusive,unintended_bias}. It has around 127,000 records with an average length of 81 words. The sensitive features include religion, country, ethnic and race. Here we assume the protected attribute is ethnic. Figure~\ref{fig:os} shows a sample $x$, note that the part ``==????=='' is the heading.
\end{example}

\subsection{Global Generation}
\label{subsec:gg}

Algorithm ~\ref{alg:gg} shows the details of the global generation phase. The algorithm uses the following constants: \emph {num\_g} which is the number of seed samples to generate during global generation; and \emph{max\_iter} which is the number of maximum iteration.

\begin{algorithm}[t]
\caption{Global Generation}
\label{alg:gg}
\begin{algorithmic}[1]
	\State id\_g = $\emptyset$
	\For{i from 0 to num\_g}
		\State Get seed $x$ from $X$ \label{li:select}
		\State Determine the locations of $P$ \label{li:prot_attr}
		\For{iter from 0 to max\_iter} \label{li:iter_begin}
			\If{$discrimination\_check(x)$} \label{li:found_begin}
				\State id\_g = id\_g $\cup$ {x}
		        \State \textbf{break} \label{li:found_end}
		    \EndIf
		    \State $X_s = \{x' |\forall x_{p}' \in \mathbb{I}_{p}, x_{p}' \ne x_{p}\}$ \label{li:potential}
		    \State $x' = \arg\max\{abs(\mathcal{O}_{y}(x')-\mathcal{O}_{y}(x)) | x' \in X_s\}$
		    \State $g = \nabla_{x} J(\theta, x, y)$
		    \State $g' = \nabla_{x'} J(\theta, x', y)$
		   	\State $x = global\_perturbation(x, grad, grad')$
		\EndFor \label{li:iter_end}
	\EndFor
	\State \Return{id\_g}
  \end{algorithmic}
\end{algorithm}

In the loop from lines~\ref{li:iter_begin} to~\ref{li:iter_end}, we generate \idie s iteratively based on the gradient. Let $\theta$ be the parameters of a DL model $\mathcal{D}$; $y$ be the ground-truth label associated with $x$; and $J(\theta, x, y)$ be the objective function, e.g., the loss function or the logits. Given a seed $x$, we first check whether it is a \idi according to Definition~\ref{de:defidi} at line~\ref{li:found_begin}. The details are shown in Algorithm~\ref{alg:dc}. Its key idea is to enumerate the value domain of the protected attributes (see lines~\ref{li:for_s}-~\ref{li:for_e}) and check whether the model labels the modified sample differently. The sample is deemed to be a \idi at line~\ref{li:dd_found} if the label changes. Note that the complexity of the checking is $\Theta(N)$, where $N$ is the number of all the possible combinations of the protected features in the corresponding domain. If $x$ is not a \idie, we start to search for \idie s based on $x$ with the guidance of the gradient defined as $\nabla_{x} J(\theta, x, y)$.

\begin{algorithm}[t]
\caption{Discrimination Check}
\label{alg:dc}
\begin{algorithmic}[1]
	\For{each element $t$ in $x$}
		\If{$t \in \mathbb{I}_{p}$}
			\For{$v$ in $\mathbb{I}_{p} \backslash \{t\}$} \label{li:for_s}
				\State Obtain $x'$ by replacing $t$ in $x$ with $v$;
				\If{$\mathcal{D}(x) \neq \mathcal{D}(x')$}  \label{li:dd_foundprior}
					\State \Return{True} \label{li:dd_found}
				\EndIf
			\EndFor \label{li:for_e}
		\EndIf
	\EndFor
	\State \Return{False}
\end{algorithmic}
\end{algorithm}

\begin{figure}[t]
\centering
\subfigure[Adversarial Attack]{
\includegraphics[width=0.33\textwidth]{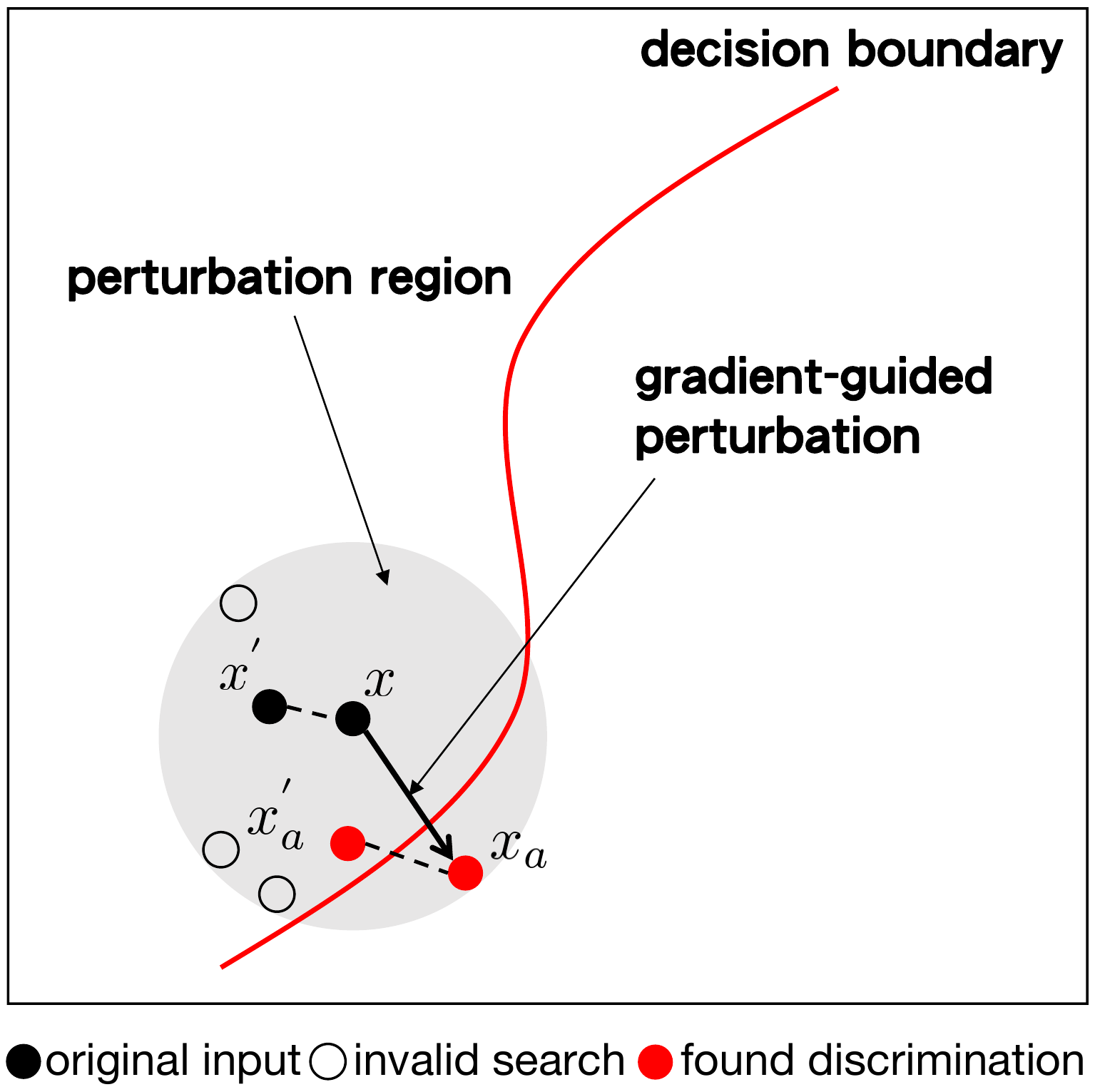}
\label{fig:adv}
}
\subfigure[Fairness Testing]{
\includegraphics[width=0.33\textwidth]{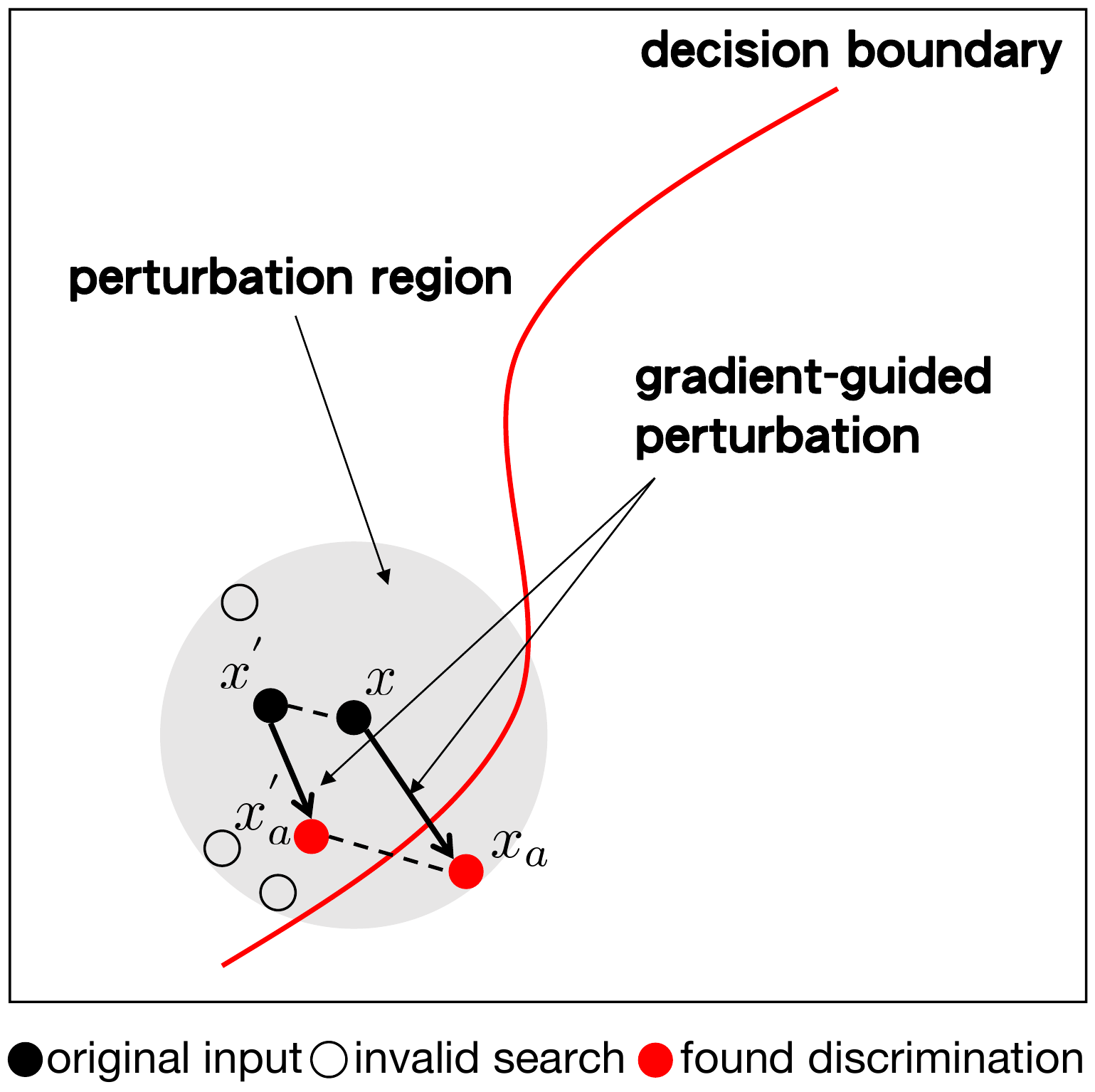}
\label{fig:ft}
}
\caption{Intuition of gradient-based approach.}
\label{fig:gradient}
\end{figure}

Notice that in order to identify a \idie, we need to find a \idi pair, i.e., a pair of samples that differ only by some protected attributes and yet have different labels. In other words, given $x$, we need to first identify an $x'$ which only differs from $x$ in protected attributes. Since $x$ is not a \idie, $x$ and $x'$ thus have the same label. There are two possible ways of perturbation to obtain \idie s. The first one (shown in Figure~\ref{fig:adv}) is directly pushing the sample $x$ towards the decision boundary by adversarial attack~\cite{fgsm,bim,jsma,hotflip,asc}, and then checking whether it is successful to acquire a \idie. In this case, if all possible $\{x'\}$ are close to $x$, they will likely cross the decision boundary together after the perturbation. Thus, the distance between $x$ and $x'$ must be considered. As shown in Figure~\ref{fig:ft}, we propose a more fine-grained strategy, which utilizes gradient information on $x$ and $x'$ simultaneously to offer guidance on how to perturb $(x,x')$ such that we are most likely to identify a \idi pair.

We identify a set of samples $X_s$ from $\mathbb{I}$ such that $x$ and any sample $x'$ in $X_s$ only differ in some protected attributes at line~\ref{li:potential}. The goal is to perturb $(x,x')$ such that $\mathcal{D}(x) \neq \mathcal{D}(x')$. Among all samples in $X_s$, we choose $x'$ according to the following equation:
\begin{equation}
x' = \arg\max \{abs(\mathcal{O}_{y}(x')-\mathcal{O}_{y}(x)) | \forall x'_{p} \in \mathbb{I}_{p}, x_{p}' \ne x_{p}\},
\end{equation}
where $\mathcal{O}$ denotes the output vector of $\mathcal{D}$. The intuition is to select the sample $x'$ such that the outputs of the DL model on $x$ and $x'$ are maximally different. In such a way, after we perturb both $x$ and $x'$, it is more likely that the predicted labels of $x$ and $x'$ are different.

Our next step is to perturb $x$ and $x'$ to generate a \idi pair $(x_{id}, x_{id}')$ such that $\mathcal{D}(x_{id})\neq\mathcal{D}(x_{id}')$. Note that the perturbation introduced to $x$ and $x'$ are always the same so as to make sure the pair still only differ by protected attributes after the perturbation. In order to minimize the perturbation, we utilize different strategies on tabular and text data since they have different characteristics. For tabular data, since the attributes are all preprocessed as categorized values, the perturbation is done by increasing or decreasing its value by 1 unit. For text data, we select the perturbation token among candidates which are similar to the original one semantically and syntactically. A \textit{perturbation} in our context is thus a function of a set of non-protected attributes which we choose to perturb and a corresponding selection vector determining the perturbing direction or token. Formally,
\begin{definition}
\textbf{\textit{Perturbation}} A perturbation $\delta$ on a data sample $x$ is function $\delta:\mathbb{I}\times NP \times \mathbb{S}\to\mathbb{I}$, where $\mathbb{S}$ is the selection of the perturbation.
\end{definition}
Specifically, for tabular data, $\mathbb{S}$ is a boolean vector where 1 means increasing the attribute value by 1 and -1 means decreasing by 1. For text data, $\mathbb{S}$ is a substituting token vector.

\begin{algorithm}[t]
\caption{Global Perturbation (Tabular)}
\label{alg:gp_tabular}
\begin{algorithmic}[1]		   	
    \State Initialize array $dir$ with the same size as $A$ by 0
    \For{$a \in A \backslash P$}
    	\If{$sign(g_{a}) = sign(g_{a}')$}
    		\State $dir_{a} = sign(g_{a})$
    	\EndIf
    \EndFor
    \State $x = x + dir * s\_g$ \label{li:perturb}
    \State $x = clip(x)$ \label{li:g_clip}
	\State \Return{x}
  \end{algorithmic}
\end{algorithm}

\begin{algorithm}[t]
\caption{Clip}{}
\label{alg:clip}
\begin{algorithmic}[1]
	\State Let $x$ be the input
	\State Let $\mathbb{I}$ be the input domain
	\For{$x_{i} \in x$}
		\State $x_{i} = \max(x_{i}, \mathbb{I}_{i}.min)$
		\State $x_{i} = \min(x_{i}, \mathbb{I}_{i}.max)$
	\EndFor
	\State \Return $x_{i}$
\end{algorithmic}
\end{algorithm}

Our next question is how to choose the elements and directions for perturbation. Notice that to better achieve individual discrimination, we need to maximize the difference between $\mathcal{D}(x_{id})$ and $\mathcal{D}(x_{id}')$ after perturbation. Our goal is thus:
\begin{equation}
argmax_{\delta(x,x')}\{\mathcal{D}(x_{id})-\mathcal{D}(x_{id}')\},
\end{equation}
where $x_{id}=\delta(x)$ and $x_{id}'=\delta(x')$. Unfortunately, this objective can not be directly optimized. Our remedy is to adopt the idea of the EM algorithm~\cite{EM} in machine learning to iteratively optimize it. Similarly, we discuss the different treatments according to the data type. For the tabular data, as showed in Algorithm~\ref{alg:gp_tabular}, we utilize the gradient of $\nabla_{x} J(\theta, x, y)-\nabla_{x} J(\theta, x', y)$ and select those attributes which have similar contributions (with the same sign of gradients) as attributes to perturb. The intuition is that perturbing these attributes can potentially enlarge the output difference since $\nabla_{x} J(\theta, x, y)-\nabla_{x} J(\theta, x', y)$ equals 0 on them (likely to be local minimum). The parameter \emph{s\_g} at line ~\ref{li:perturb} is the step size of global generation. Notice that in order to filter out unreal perturbed inputs, we always apply a Clip function described in Algorithm ~\ref{alg:clip} to make sure that the value of each attribute after the perturbation is within its domain (see line ~\ref{li:g_clip}).

\begin{example}
\label{exp:1.2}
For our tabular running example, we select seed sample from the original dataset, then get the first seed $x$ is as follows (shown in Example~\ref{exp:1.1}), and it is not a \idie.
$$x: [4, 0, 6, 6, 0, 1, 2, 1, \textcolor{red}{1}, 0, 0, 40, 100]$$
We identify all samples which differ from the seed only by protected attributes and then obtain the following $x'$ which has the greatest difference in output probability with $x$.
$$x': [4, 0, 6, 6, 0, 1, 2, 1, \textcolor{red}{0}, 0, 0, 40, 100]$$
We then determine the perturbation direction based on the sign of two samples' gradients as follows.
$$direction: [0, 1, 0, 0, -1, 1, -1, 0, 1, 0, 0, 1, -1]$$
\noindent
Intuitively, 0 means that the corresponding attribute should not be changed; $-1$ means that it should be decreased and $1$ means that it should be increased (to maximize output difference). Next, we perturb $x$ accordingly and apply the Clip function to filter invalid values. The result is the following sample.
$$x: [4, 1, 6, 6, 0, 2, 1, 1, \textcolor{red}{1}, 0, 0, 41, 39]$$
\noindent Since the last attribute \emph{native-country} only has 40 countries (and value 100 means missing value in the original data), it is modified to 39 (the maximum value) by the Clip function. After checking at line~\ref{li:found_begin}, it is shown to be a \idie.
\end{example}

\begin{algorithm}[t]
\caption{Global Perturbation (Text, ASC)}
\label{alg:gp_text1}
\begin{algorithmic}[1]
    \State $a = \arg\max_{a \in \{A \backslash P\}} imp(a)$\label{li:impactful}
    \State $C = x_a.simlar\_words$ \label{li:candidate}
    \State $x_a = \arg\min_{c \in C}\|sign(x_{a}-c)-sign(g_{a})\|$
	\State \Return{x}
  \end{algorithmic}
\end{algorithm}

\begin{algorithm}[t]
\caption{Global Perturbation (Text, HF)}
\label{alg:gp_text2}
\begin{algorithmic}[1]
	\State $min\_dif = 0$
	\For{$a \in \{A \backslash P\}$} \label{li:search_begin}
		\State $C = x_a.simlar\_words$ 
		\For{$c \in C$}
			\State $dif = |grad_{a} \cdot (x_a - c)^T - grad'_{a} \cdot (x'_a - c)^T|$ \label{li:loss_change}
			\If{$dif > min\_dif$}
				\State $pt = a, sw = c$
			\EndIf
		\EndFor
	\State $x_{pt} = sw$
	\EndFor \label{li:search_end}
	\State \Return{x}
\end{algorithmic}
\end{algorithm}

For models trained on text data, we propose two different global generation strategies, both of which are based on gradient. The first one is inspired by ASC~\cite{asc} (shown in Algorithm~\ref{alg:gp_text1}), i.e., to maintain the semantics and syntax, we only choose the most impactful token to perturb (see line~\ref{li:impactful}), instead of perturbing multiple attributes at the same time. Recall that the tokens in the text are pre-processed by word embedding, which projects each token to a numerical vector, i.e., the gradient is calculated with respect to the projected embedding vectors (instead of the token directly). Formally, the impact of a non-sensitive token $np$ is measured by
\begin{equation}
\label{eq:imp}
imp(np)= ||abs(g_{np}-g'_{np})||_{\infty}
\end{equation}
where $abs(\cdot)$ returns the absolute value of $\cdot$; and $||\cdot||_{\infty}$ is the $L_\infty$ norm of a vector $\cdot$. After getting the most impactful token, we replace it with one of its synonyms at line~\ref{li:candidate}. Among the top-$k$ synonyms, denoted as $C$, we select the one which incurs a perturbation whose direction is closest to the direction indicated by the gradient of $x$. The intuition is to push the sample towards the classification boundary along the gradient so that it is more likely to be a \idie. 

The other method inherits the key idea of HotFlip~\cite{hotflip}, as shown in Algorithm~\ref{alg:gp_text2}. We substitute each word (see lines~\ref{li:search_begin}-~\ref{li:search_end}) to search for the maximum difference in terms of the outcome change with respect to the input pair $x$ and $x'$ (see line~\ref{li:loss_change}). Since we use word embedding to preprocess the input, instead of one-hot representation, it is inapplicable to choose the best word flip $w_i \to w_j$ based on $grad_{j} - grad_{i}$ (refer readers to~\cite{hotflip} for more details), we utilize $grad_{a} \cdot (x_a - c)^T$ to estimate the change of prediction, where $\cdot^T$ is the transpose function. Notice that since here we only consider the non-protected tokens, $x'_a=x_a$.

Once we determine the perturbation, we apply it on $x$ and $x'$ and then check whether the new pair $(x_{id},x_{id}')$ is a \idi pair. If the answer is yes, the algorithm breaks out immediately (see lines~\ref{li:found_begin}-\ref{li:found_end}). Otherwise, we start another round of perturbation on $(x_{id},x_{id}')$. This process may repeat multiple times until it succeeds. 

\begin{example}
\label{exp:2.2}
For our text running example shown in Figure~\ref{fig:os}, it is also determined to be a non-\idie . We thus apply global generation (Algorithm~\ref{alg:gp_text1}) to acquire a \idi based on $x$. We first enumerate all samples by systematically replacing the sensitive word ``African'' with all other possible values for ethnicity. Among all texts, we select $x'$ where `African' is replaced by `Latino' whose prediction result is maximally different from the original text. Then we identify the non-sensitive token ``Jews'' (highlighted in red) which has the most impact on the prediction result according to the gradients. Next, we replace ``Jews'' with the top-10 synonyms in the given dictionary, which are \emph{Jewish, Christians, Nazis, Catholics, Muslims, Arabs, Jew, Germans, Gypsies, and Orthodox}. ``Gypsies'' is chosen to replace ``Jews'' according to the above discussion. The resultant text is shown in Figure~\ref{fig:gs} is a \idie .
\end{example}

\subsection{Local Generation}
\label{subsec:lg}
After the global generation phase, we obtain a set of \idie s as seeds for the local generation phase. The goal of the local generation phase is to generate as many \idie s as possible based on the seeds, which are useful for re-training the DL model. The intuition behind the design of the local generation is that a well-trained DL model is likely robust, i.e., if two samples are similar, the same prediction is likely to be produced by the DL model. We thus are likely to find more \idie s around a given seed \idie.

\begin{algorithm}[t]
\caption{Local Generation}
\label{alg:lg}
\begin{algorithmic}[1]
	\State id\_l = $\emptyset$
	\For{$x \in id\_g$}
		\For{i from 0 to num\_l} \label{li:another}
			\State $X_s = \{x^{'} |\forall x^{'} _{p} \in \mathbb{I}_{p}, x^{'}_{p} \ne x_{p}\}$
			\State $\exists x^{'} \in X_s, \mathcal{D}(x) \ne \mathcal{D}(x^{'})$ \label{li:pair}
			\State $g = \nabla_{x} J(\theta, x, y)$ \label{li:loss1}
		    \State $g' = \nabla_{x'} J(\theta, x', y)$ \label{li:loss2}
		    \State $prob = normalization(g, g^{'})$ \label{li:norm}
		    \State Select $a \in A \backslash P$ with probability $prob$
		    \State x = local\_perturbation(x, a)	
		    \If{$discrimination\_check(x)$} \label{li:l_check}
				\State id\_l = id\_l $\cup$ {x}
		    \EndIf
		\EndFor
	\EndFor
	\State \Return{l\_id}
  \end{algorithmic}
\end{algorithm}

Our local generation algorithm has the following parameters: \emph{num\_l} which is the number of trials in the local generation. The algorithm makes use of the gradients of the objective function (see lines~\ref{li:loss1}-\ref{li:loss2}) in a different way. Recall that in the global generation, we would like to maximally change the output of the DL model. On the contrary, in the local generation, we would like to minimally change the output, as our goal is to maintain the DL model's outputs of the \idi pairs identified in the global generation so that they remain different. We thus choose to perturb those attributes which have the least impact on the output. Note that the absolute value of gradient represents how much an attribute contributes to the outcome.

\begin{algorithm}[t]
\caption{Normalization}
\label{alg:normalize}
\begin{algorithmic}[1]
    \State Initialize gradient with the same size of g
    \For{i from 0 to gradient.length}
    	\State $saliency = \|abs(g_{i})\|_{\infty} + \|abs(g_{i}')\|_{\infty}$ \label{li:saliency}
    	\State $gradient_{i} = 1.0 / saliency$ \label{li:reciprocal}
    	\If{$A_{i} \in P$} \label{li:filter_begin}
    		\State $gradient_{i}=0$
    	\EndIf \label{li:filter_end}
    \EndFor
    \State $gradient\_sum = sum(gradient)$ \label{li:nor_begin}
    \State $probability = \{gd / gradient\_sum | \forall gd \in gradient\}$ \label{li:nor_end}
    \State \Return{probability}
\end{algorithmic}
\end{algorithm}

Further note that since we are perturbing a \idi pair, we need to consider the two inputs at the same time (to make sure that neither of them crosses the decision boundary as otherwise, they are no longer a \idi pair). To achieve that, we adopt a normalization process on the gradients on the two inputs to measure the average contribution of each attribute on the input pair. The details are shown in Algorithm~\ref{alg:normalize}. We first add the absolute value of two gradients together to get the saliency value of each element (see line~\ref{li:saliency}). Recall that in the case of text data, the gradient with regard to the token is a vector, thus we take the $L_{\infty}$ norm of it as impact. Then we calculate the reciprocal value (see line~\ref{li:reciprocal}) since we aim to select the element with fewer contributions to the output and meanwhile filter out the protected attributes (see lines~\ref{li:filter_begin}-\ref{li:filter_end}). Lastly, we use a standard normalization function to get the contribution of each attribute on the input pair (see lines~\ref{li:nor_begin}-\ref{li:nor_end}).

\begin{algorithm}[t]
\caption{Local Perturbation (Tabular)}
\label{alg:lp_tabular}
\begin{algorithmic}[1]	
	\State Select $d \in [1,-1]$ with probability $[0.5,0.5]$
	\State $x_{a} = x_{a} + d \times s\_l$
	\State $x = clip(x)$ \label{li:l_clip}	
	\State \Return{x}
  \end{algorithmic}
\end{algorithm}

\begin{algorithm}[t]
\caption{Local Perturbation (Text)}
\label{alg:lp_text}
\begin{algorithmic}[1]		   	
    \State $C = x_a.simlar\_words$ \label{li:candidate}
    \State Select $c \in C$ with the same probability $1.0/C.length$
    \State $x_a = c$
	\State \Return{x}
  \end{algorithmic}
\end{algorithm}

Algorithm~\ref{alg:lg} shows the details of our local generation algorithm. Given a \idi pair $(x,x')$ (see line ~\ref{li:pair}), we start searching by iteratively selecting the attribute to perturb using the normalization of gradients (see line ~\ref{li:norm}). Instead of modifying the chosen element based on the gradient, we randomly select the perturbation direction (see Algorithm~\ref{alg:lp_tabular}) and substituting word (see Algorithm~\ref{alg:lp_text}) for tabular and text data respectively with the uniform probability. Similar to global generation, we also utilize application-specific strategies (i.e., Clip for tabular and similar words for text) to make sure that the generated test case is valid. We check whether the input after the perturbation is a \idi (see line ~\ref{li:l_check}) and continue to the next seed input if the answer is yes. Otherwise, we start another iteration of the local generation (see line ~\ref{li:another}).

\begin{example}
\label{exp:1.3}
For our tabular running example, the global generation phase generates the following \idi pair.
$$x: [4, 1, 6, 6, 0, 2, 1, 1, \textcolor{red}{1}, 0, 0, 41, 39]$$
$$x': [4, 1, 6, 6, 0, 2, 1, 1, \textcolor{red}{0}, 0, 0, 41, 39]$$
\noindent In the local generation, taking this pair as input, we first calculate the gradient of these two samples and normalize the sum of them as individual probability. The result is as follows.
$$probablity: [0.030, 0.019, 0.057, 0.075, 0.002, 0.009, $$
$$~~~~~0.020, 0.015, 0, 0.002, 0.027, 0.612, 0.131]$$
\noindent Based on the above probability, we choose the attribute \emph{hours-per-week} (with index 11) and the direction -1 for perturbation. Since the result sample's values are all within the respective domains, the Clip function keeps it the same and the following sample is identified as a new \idie.
$$x: [4, 1, 6, 6, 0, 2, 1, 1, \textcolor{red}{1}, 0, 0, 40, 39]$$
\end{example}

\begin{example}
\label{exp:2.3}
For our text running example, we apply local generation to search for more \idie s on the neighbor of the identified one (shown in Figure~\ref{fig:gs}). We first obtain the probability distribution based on the normalization of their gradients and randomly select the token ``many'' to perturb. Note that this token has little impact on the prediction result. Next, we obtain its top-10 synonyms and randomly replace ``many'' using ``several''. The resultant text is shown in Figure~\ref{fig:ls}, which turns out to be a \idie .
\end{example}

\subsection{Qualitative Evaluation} 
\label{qualitative}
In the following, we evaluate our approach qualitatively by comparing it with state-of-the-art approaches, i.e., \tm~\cite{themis}, \aeq~\cite{aequitas}, and Symbolic Generation (SG)~\cite{sg}. Empirical comparison results are presented in Section~\ref{sec:experiment}.

\tm~\cite{themis} explores the input domains for all attributes through random sampling and then checks whether the generated samples are \idie s. \aeq~\cite{aequitas} improves \tm by adopting a two-phase generation framework. In the first phase, \aeq randomly generates a set of \idie s in the input space as seeds. In the second phase, \aeq searches for more \idie s around the seed inputs found in the first phase by randomly adding perturbations on the non-protected attributes. Notice that the perturbation is guided by a distribution that describes the probability of finding a \idi by adding perturbation on a specific non-protected attribute. Despite that random sampling is lightweight, \tm and AEQUITAS can miss many combinations of non-protected attributes' values where individual discrimination may exist~\cite{sg}. The recent work SG~\cite{sg} attempts to solve this problem by systematically exploring the input space through symbolic execution. The idea is to first adopt a local model explainer like LIME~\cite{lime} to construct a decision tree for approximating the machine learning model. The result is a decision tree constituted with linear constraints such that a linear path constraint is associated with any given input. Then, SG iteratively selects (according to a ranking function), negates the constraints, and uses a symbolic execution solver to generate test cases according to different path constraints.

To summarize the difference between existing approaches and ours, we differentiate them using three criteria, i.e., whether the search (for \idie s) is guided, whether the guided search is specific for an individual input (input-specific), and whether the procedure adopted is light-weight (and thus likely scalable). Table~\ref{tab:feature} shows the summary. Except for \tme, both \aeq and SG generate \idie s in a guided way (either by a distribution or a path constraint). The difference is that \aeq uses a single distribution for all the inputs while SG generates path constraints depending on different inputs. We remark that designing input-specific perturbations is a more robust way to generate individual discrimination for different kinds of input and thus is important because it is crucial for removing individual discrimination globally. Lastly, we expect that approaches based on random sampling like \tm and \aeq are lightweight while SG is a relatively heavy approach that requires the help of a local model explainer and a symbolic execution solver. For the former, it is still an open problem on generating model explainers in a scalable and accurate way. For the latter, symbolic execution is known to be less scalable than techniques like random samples. Besides, the above limitations also make these methods unsuitable for text application since it is challenging to obtain valid texts by randomly selecting words. For SG, it is impractical to solve the constraint after the embedding procedure. 

\begin{table}[t]
\centering
\caption{Comparing different approaches}
\label{tab:feature}
\begin{tabular}{|c|c|c|c|c|}
\hline
Feature & \tm & \aeq & SG & \method \\
\hline
Guided          & \no       & \yes (semi)         &  \yes  & \yes    \\
Input specific  & N.A.       & \no         &  \yes  & \yes    \\
Lightweight & \yes       & \yes         &  \no  & \yes    \\ 
\hline
\end{tabular}
\end{table}

Compared to existing approaches, our approach satisfies all three criteria. First, our search is guided by gradient, i.e., the perturbation is guided towards the decision boundary to accelerate the discovery of \idie s which significantly reduces the number of attempts needed. The intuition is visualized in Figure~\ref{fig:gradient}. Second, our algorithm generates a specific gradient-guided search for different inputs, which significantly improves the success rate of individual discrimination generation. Lastly, our approach is lightweight since obtaining the gradient of DL models with respect to a given input is cheap which only requires a backpropagation process and is supported by all existing standard deep learning frameworks like Tensorflow~\cite{tensorflow}, PyTorch, and Keras. And \method is easy to be implemented on text data, as no matter on global or local generation phase, it selects the perturbed token and substitutes word only guided by gradients.

Similar to \aeqe, our approach also has a global search phase and a local search phase. The differences are in the details of both phases. \aeq works by actively maintaining a probability distribution $t:NP\to[0,1]$ on $NP$ which represents how likely perturbing an attribute in $NP$ is likely to successfully generate individual discriminatory samples. A limitation of such an approach is that different attributes of different inputs may contribute differently to the DNN output and the same global distribution hardly works for all the inputs. This is clearly evidenced by our experiment results in Section~\ref{sec:experiment}. To solve the problem, our approach takes an input-specific perspective, i.e., choosing different local perturbations based on the gradient which is specific to a given sample.

%% file: 4_Experiment.tex
\section{Experiment}
\label{sec:experiment}
We have implemented \method as a self-contained toolkit based on Tensorflow ~\cite{tensorflow} and Gensim~\cite{gensim}. Its source code, together with all the experiment-related details, is available online~\cite{appendix}. In the following, we evaluate \method to answer \questions research questions (RQ). 

\subsection{Experimental Setup}
\label{subsec:setup}
For the tabular data, we choose \aeq and SG for baseline comparison. Note that \tm is shown to be significantly less effective~\cite{sg} and thus is omitted for comparison. We obtained the implementation of \aeq from GitHub\footnote{https://github.com/sakshiudeshi/Aequitas} and re-implemented SG according to the description in~\cite{sg} since their implementation is not publicly available. Further notice that \aeq proposed 3 different local search algorithms, we adopted the fully-directed algorithm in our evaluation since it has the best performance according to~\cite{aequitas}. However, for text classification tasks, as all existing fairness testing approaches are specifically designed on tabular data and thus inapplicable, we compare \method with a baseline approach which adopts the key idea of \tm and \aeqe , applying random perturbation (RP) instead of gradient-guided perturbation (to show the effectiveness of gradient-guided search). In particular, both on the global and local stage, RP selects the perturbed token and substitutes one with synonyms completely at random.

\begin{table}[t]
\centering
\caption{Configuration of experiments.}
\label{tab:configuration}
\begin{tabular}{|c|c|c|}
\hline
Parameter & Value & Description\\
\hline
max\_iter & 10 & max. iteration of global generation \\
s\_g & 1.0 & step size of global generation \\
s\_l & 1.0 & step size of local generation \\
k & 10 & the number of similar words for choosing \\
\hline
\end{tabular}
\end{table}

All experiments are conducted on a GPU server with 1 Intel Xeon 3.50GHz CPU, 64GB system memory, and 1 NVIDIA GTX 1080Ti GPU. Both \aeq and SG are configured according to the best performance setting reported in the respective papers. Table~\ref{tab:configuration} shows the value of parameters used in our experiment to run \methode .

We adopt \datasets open-source datasets for fairness testing as our experiment subjects, including \tabulardata tabular datasets and \textdata text datasets. The details of the datasets are as follows.
\begin{itemize}
	\item \emph{Census Income} The details of this dataset have been introduced in Example~\ref{exp:1.1}. It is used as a benchmark by \aeq and SG~\cite{sg}.
	\item \emph{German Credit}\footnote{https://archive.ics.uci.edu/ml/datasets/statlog+(german+credit+ data)} This is a small dataset with 600 data and 20 attributes. It was used to evaluate several existing works~\cite{themis,sg}. The attributes of \emph{age} and \emph{gender} are protected attributes. The original aim of the dataset is to give an assessment of an individual's credit based on personal and financial records. It is used as a benchmark by SG~\cite{sg}.
	\item \emph{Bank Marketing}\footnote{https://archive.ics.uci.edu/ml/datasets/bank+marketing} The dataset came from a Portuguese banking institution and is used to train models for predicting whether the client would subscribe a term deposit based on his/her information. The size of the dataset is more than 45,000. There are a total of 16 attributes and the only protected attribute is \emph{age}. It is used as a benchmark by SG~\cite{sg}.
	\item \emph{Wikipedia Comments} The overview of this dataset has been summarized in Example~\ref{exp:2.1}.
	\item \emph{Jigsaw Comments}\footnote{https://www.kaggle.com/c/jigsaw-toxic-comment-classification-challenge} This is a public dataset from an online competition that was held by Jigsaw and Google. It is also a commonly used benchmark for fairness research~\cite{counterfactual}. It has around 313,000 comments with an average length of 80 words classified into six categories of toxicity (i.e., toxic, severe toxic, obscene, threat, insult, and identity hate) and non-toxic. The sensitive features are the same as those for the Wikipedia Comments.
\end{itemize}

For the tabular data, we use the binning method to pre-process the numerical attributes. For the text data, in order to balance accuracy and efficiency, we adopt the state-of-art embedding tool GloVe~\cite{glove} and use the 50-dimension pre-trained word vectors\footnote{http://nlp.stanford.edu/projects/glove} trained on 6 billion tokens and 400 thousand vocabularies of Wikipedia 2014 and Gigaword 5. For those out-of-vocabulary words, we take the average of all the word vectors as an unknown vector suggested by the author of GloVe.

\begin{figure}[t]
\centering
\includegraphics[scale=0.5]{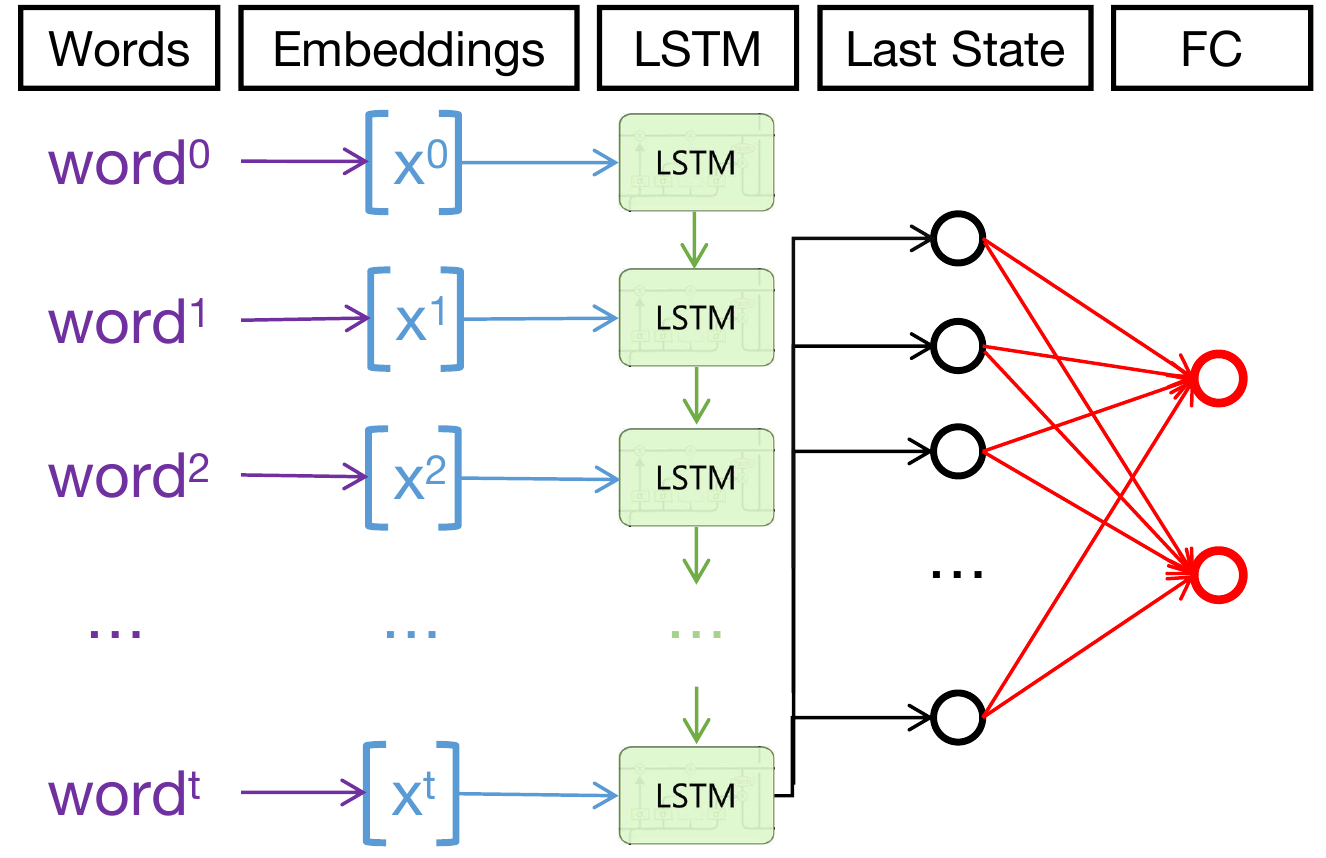}
\caption{Experimental LSTM-based model.}
\label{fig:lstm}
\end{figure}

\begin{table}[t]
\centering
\caption{Experimented DL models.}
\label{tab:acc}
\begin{tabular}{|c|c|c|}
\hline
Dataset & Model & Accuracy \\
\hline
Census Income & Six-layer Fully-connected NN & 88.15\% \\
German Credit & Six-layer Fully-connected NN & 100\%  \\
Bank Marketing & Six-layer Fully-connected NN & 92.26\% \\
\hline
\multirow{2}{*}{Wiki Comments} & 3-layer 10-state LSTM & 89.9\% \\
 & 3-layer 10-state GRU & 90.7\% \\ \hline
\multirow{2}{*}{Jigsaw Comments} & 3-layer 10-state LSTM & 85.8\% \\
 & 3-layer 10-state GRU & 86.8\% \\
\hline
\end{tabular}
\end{table}

The details of the models used in the experiments are shown in Table~\ref{tab:acc}. Since the experimented tabular datasets are relatively simple, we train models in the form of fully-connected DNN. We adopt two common improved variants of RNN, Long Short Term Memory (LSTM)~\cite{lstm} and Gated Recurrent Unit (GRU)~\cite{gru}, both with 3 layers and 10 hidden states, on the text datasets. After getting the final state of the LSTM and GRU models, we apply a fully connected layer to output the predicted labels. Figure~\ref{fig:lstm} shows the overview of the LSTM-based tested model.

The key KPIs for the comparison between our algorithm and baselines are the number of \idie s, the success rate of generating \idie s, and the generation efficiency. For text data, we additionally compare the quality of generated discriminatory text.

\subsection{Research Questions}
\label{subsec:rq}
We aim to answer the following research questions through our experiments. \\

\label{subsec:questions}
\noindent \emph{RQ1: How effective is our algorithm in finding \idie s?}

\noindent \textbf{Tabular data} We first compare \method with \aeq and SG on tabular datasets. Since AEQUITAS and \method both have a global generation phase and a local generation phase, we conduct a detailed comparison for both phases. For both of them, we generate 1,000 samples in the global generation phase (except for credit data, which is set to be 600 due to its small size), and then generate 1,000 samples during local generation for each successfully identified \idi in the global phase.

\begin{table}[t]
\centering
\caption{Comparison with \aeqe. \#GDiff denotes the number of unique generated samples, and \#ID denotes the number of identified \idie s.}
\label{tab:comparison_aequitas}
\resizebox{.48\textwidth}{!}{
\begin{tabular}{|c|c|c|c|c|c|}
\hline
\multirow{2}{*}{Dataset} & \multirow{2}{*}{Prot. Attr.} & \multicolumn{2}{c|}{AEQUITAS} & \multicolumn{2}{c|}{\method} \\
\cline{3-6} & & \#GDiff & \#ID & \#GDiff & \#ID \\
\hline
census & age & 56955 & 6045 & 491650 & 256980\\
census & race & 54734 & 4737 & 344162 & 139179\\
census & gender & 32148 & 2930 & 259227 & 47644\\
bank & age & 32870 & 8949 & 700285 & 364758\\
credit & age & 99560 & 38479 & 398209 & 233664\\
credit & gender & 33137 & 4996 & 312919 & 73497\\
\hline
\end{tabular}
}
\end{table}

Table~\ref{tab:comparison_aequitas} presents the details of the comparison. Note that the maximum number of the two-stage searched samples is thus 1,001,000 (i.e., 1,000 global and 1,000*1,000 local samples). We further filter out the duplicate samples. Column \#GDiff is the number of non-duplicate samples generated after the two-phase search. Column \#ID shows the number of \idie s identified. It can be observed that \method is significantly more effective than AEQUITAS in finding \idie s. \textit{On average, \method generates 8.6 times more non-duplicate samples.} One of the reasons why AEQUITAS explores a much smaller space is that it often generates duplicate samples (as was observed in~\cite{sg}) since a global sampling distribution is used for all the inputs, while \method perturbs a specific input according to the guidance of an sample-specific gradient. More importantly, \textit{\method generated nearly 25 times more \idie s on average.} A close investigation shows that the reason is that gradient provides much better guidance in identifying \idie s. This is evidenced by the average success rate which is calculated by \#ID $/$ \#GDiff. \textit{AEQUITAS has a success rate of 18.22\%, whereas \method achieves a success rate of 40.89\%, which is more than 2 times that of AEQUITAS.}

\begin{table}[t]
\centering
\caption{Comparison with SG in 500 seconds. \#GDiff denotes the number of unique generated samples, and \#ID denotes the number of identified \idie s.}
\label{tab:comparison_sg}
\resizebox{.48\textwidth}{!}{
\begin{tabular}{|c|c|c|c|c|c|}
\hline
\multirow{2}{*}{Dataset} & \multirow{2}{*}{Prot. Attr.} & \multicolumn{2}{c|}{SG} & \multicolumn{2}{c|}{\method} \\
\cline{3-6} & & \#GDiff & \#ID & \#GDiff & \#ID \\
\hline
census & age & 1290 & 544 & 8202 & 3453\\
census & race & 1541 & 632 & 10677 & 4290\\
census & gender & 1482 & 280 & 22977 & 4164\\
bank & age & 1385 & 842 & 5911 & 3587\\
credit & age & 2752 & 1574 & 5771 & 3923\\
credit & gender & 3202 & 926 & 14711 & 4091\\
\hline
\end{tabular}
}
\end{table}

Although SG similarly has two phases, it works differently from \method or AEQUITAS. That is, SG maintains a priority queue, pops a sample, and applies global search iteratively. If the sample is a \idie, the local search is employed. Afterward, all the search results are pushed into the queue without checking whether they are discriminatory or not. As a result, it is infeasible to directly compare SG and \method as above. Thus, we apply an overall evaluation between \method and SG within the same time limit, i.e., 500 seconds. The results are shown in Table ~\ref{tab:comparison_sg}. We observe that on average: \textit{\method 1) explores 6.6 times more samples, 2) generates 6.5 times more \idie s, and 3) has a 42.8\% success rate (whereas SG has a success rate of 41.5\%).} One thing to notice is that our method beats SG which is based on a symbolic solver even in terms of success rate. One possible explanation is that the model explainer SG utilized is far from accurate for complex models like DL models. 

In addition to the above overall evaluation with two baselines, we further conduct a comprehensive comparison phase by phase, i.e., global generation and local generation.

\begin{table}[t]
\centering
\caption{Number of \idie s generated by global generation.}
\label{tab:g_comparison}
\begin{tabular}{|c|c|c|c|c|}
\hline
Dataset & Protected Attr. & AEQUITAS & SG & \method \\
\hline
census & age & 101 & 291 & 658\\
census & race & 95 & 139 & 456\\
census & gender & 37 & 54 & 334\\
bank & age & 43 & 142 & 872\\
credit & age & 175 & 247 & 594\\
credit & gender & 47 & 87 & 451\\
\hline
\end{tabular}
\end{table}

\vbox{}
\noindent \emph{Global generation} The goal of the global generation is to identify diversified \idie s. For a fair comparison, we generate 1,000 samples in the global phase (except for \emph{credit data}, which is set to be 600), and count how many \idie s are identified by each method in this stage. Note that the same seed samples are used for SG and \methode. 

The results are shown in Table~\ref{tab:g_comparison}. It can be observed that \method generates the most number of \idie s, \textit{with an average improvement of 794\% and 324\% when it is compared with AEQUITAS and SG respectively.} We take that this shows the effectiveness of guiding the search based on gradient during global generation.

\begin{table}[t]
\centering
\caption{Number of \idie s generated by local generation.}
\label{tab:l_comparison}
\begin{tabular}{|c|c|c|c|c|}
\hline
Dataset & Protected Attr. & AEQUITAS & SG & \method \\
\hline
census & age & 216 & 422 & 598\\
census & race & 153 & 371 & 526\\
census & gender & 189 & 210 & 321\\
bank & age & 221 & 634 & 708\\
credit & age & 448 & 600 & 750\\
credit & gender & 142 & 280 & 337\\
\hline
\end{tabular}
\end{table}

\vbox{}
\noindent \emph{Local generation} The local generation aims to further craft more \idie s based on the results of the global generation. To make a fair comparison between the three strategies for local generation, we seed each method with the same set of \idie s and apply the three strategies to generate 1,000 samples for each seed. In this way, we are able to properly evaluate the local generation strategies without being influenced by the results of the global generation adopted by the three methods. 

Table~\ref{tab:l_comparison} shows the results of the comparison. It is obvious that the local generation strategy of our method \method performs the best among the three. Specifically, \textit{ADF generates 153\% more \idie s than AEQUITAS, and 32\% more than SG on average.}

\begin{figure}[t]
\centering
AEQUITAS
\includegraphics[width=0.45\textwidth]{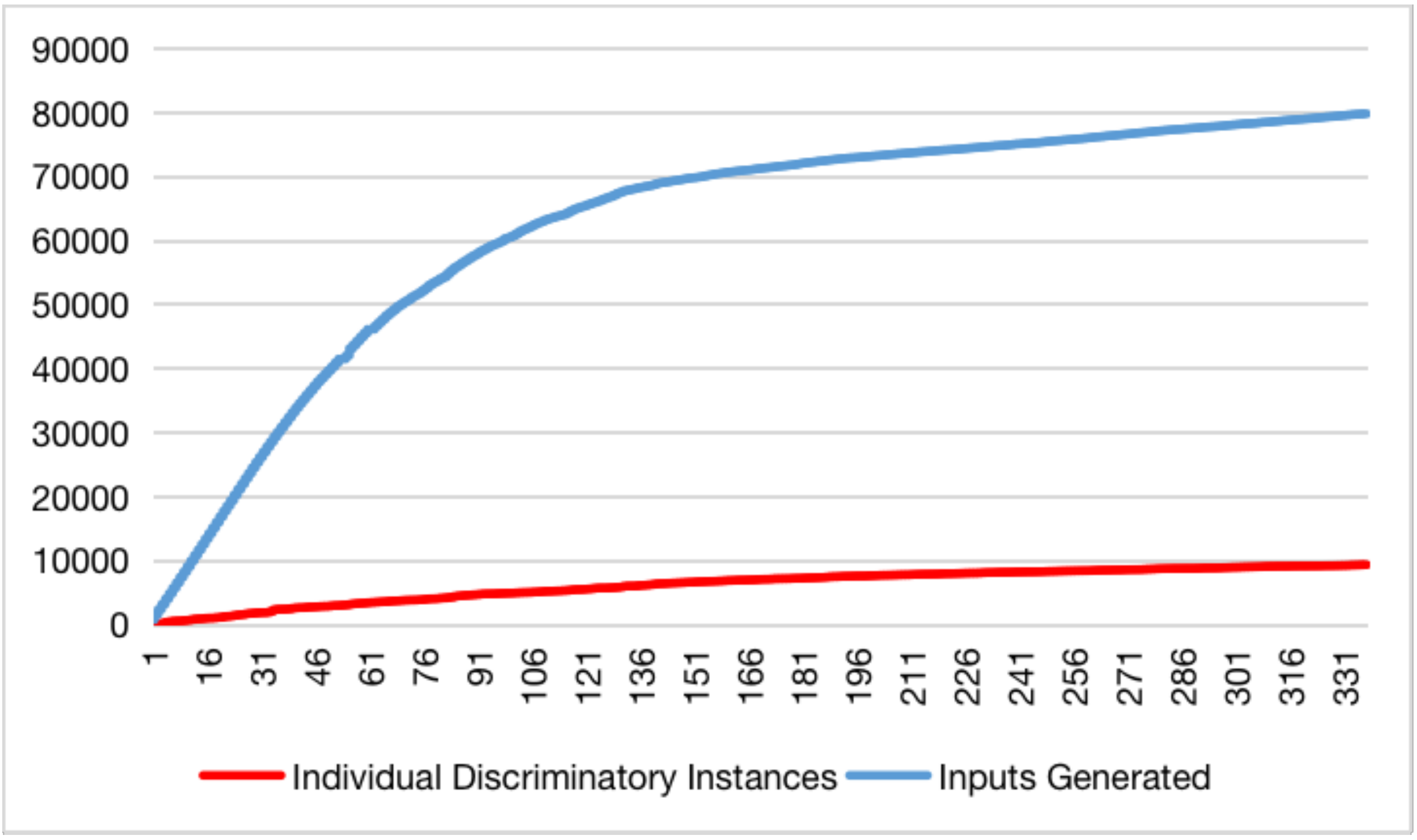}
ADF
\includegraphics[width=0.45\textwidth]{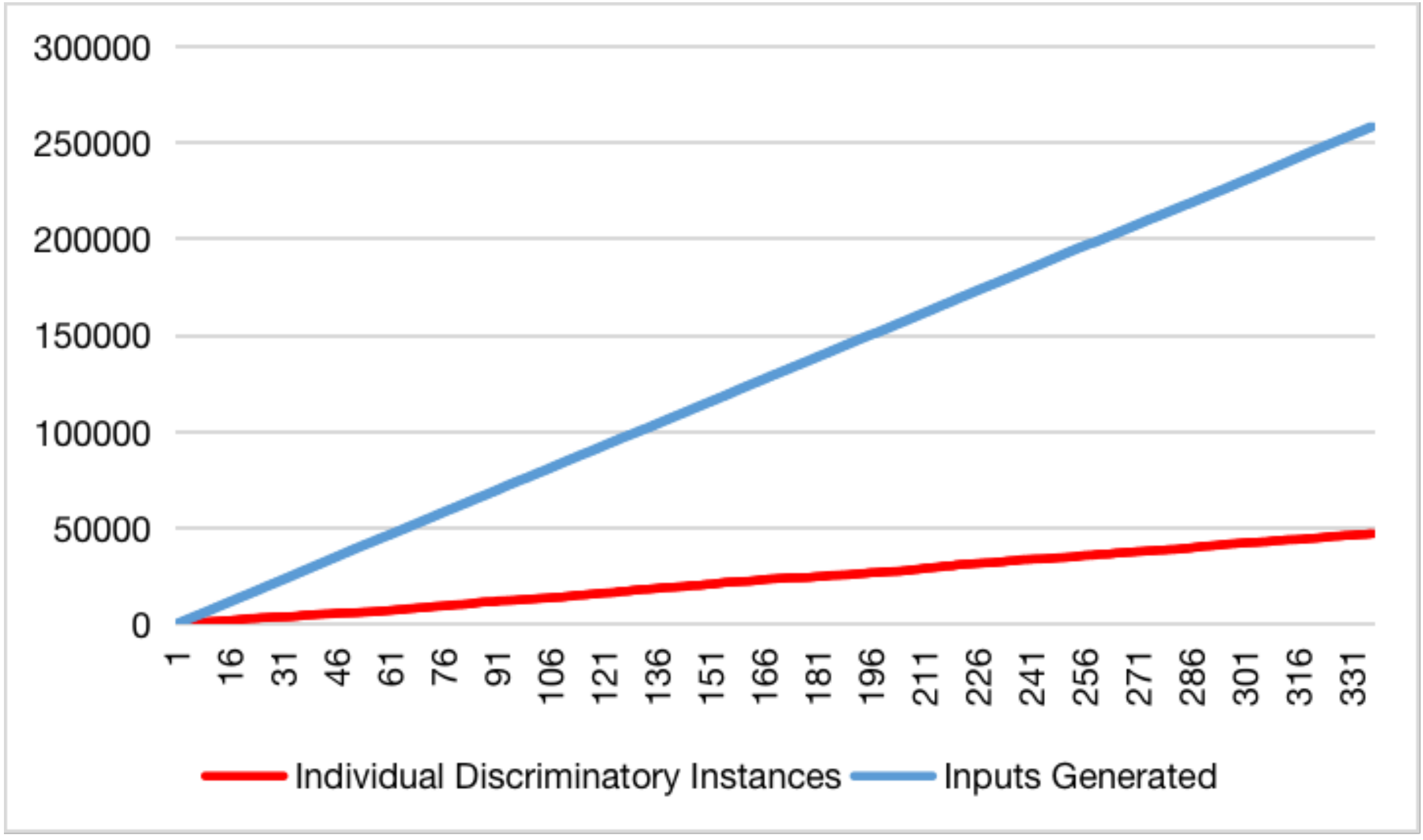}
\caption{Effectiveness of local generation.}
\label{fig:ADF_local}
\end{figure}

Recall that AEQUITAS and ADF both guide local generation through a probability distribution which intuitively is the likelihood of identifying \idie s by changing certain attributes. The difference is that AEQUITAS's probability is global, i.e., the same probability is used for all samples, whereas ADF's probability is based on gradient and thus specific to the certain sample. We conduct a further experiment to evaluate whether ADF's approach is more effective or not. We feed these approaches the same set of seed samples and then measure the relationship between the number of seed samples explored and the number of new \idie s identified.

The result is shown in Figure~\ref{fig:ADF_local} where the x-axis is the number of seeds explored; the blue line represents the total number of samples generated and the red line represents the number of \idie s identified. It can be observed that for \methode , both lines grow steadily with the number of seeds explored, which suggests that the sample-specific probability used in ADF works reliably. In comparison, the increase of both the number of samples and the number of \idie s drops with an increasing number of seeds for AEQUITAS. This is possibly due to the ineffective global probability and the duplication in the generated samples.

\begin{table*}[t]
\centering
\caption{Effectiveness. }
\label{tab:effective}
\begin{tabular}{|c|c|c||c|c|c|c|c|c|c||c|c||}
\hline
Dataset & Model & Protected Attr. & \multicolumn{7}{|c||}{Global Generation} & \multicolumn{2}{|c||}{Local Generation} \\ \cline{4-12}
& & & Seed & Raw & RP & ASC & \methode\_ASC & HF & \methode\_HF & RP & \method \\
\hline
\multirow{8}{*}{Wiki} & \multirow{4}{*}{LSTM} & country & 1000 & 31 & 29 & 119 & 131 & 308 & 401 & 47094 (80.6\%) & 145481 (92.1\%)\\
& & ethnic & 1000 & 117 & 29 & 123 & 166 & 326 & 423 & 128174 (90.6\%) & 261746 (94.9\%)\\
& & race & 450 & 24 & 17 & 76 & 95 & 157 & 177 & 32967 (83.0\%) & 107678 (92.8\%)\\
& & religion & 358 & 29 & 14 & 50 & 64 & 117 & 150 & 37382 (88.7\%) & 86350 (94.6\%)\\ \cline{2-12}
& \multirow{4}{*}{GRU} & country & 1000 & 33 & 22 & 83 & 100 & 303 & 405 & 43580 (81.5\%) & 117541 (90.9\%)\\
& & ethnic & 1000 & 52 & 34 & 121 & 142 & 366 & 429 & 70093 (84.7\%) & 174488(92.4\%)\\
& & race & 450 & 27 & 11 & 56 & 82 & 172 & 208 & 30651 (82.8\%) & 93944 (88.2\%)\\
& & religion & 358 & 19 & 9 & 31 & 39 & 122 & 143 & 24748 (90.3\%) & 53866 (94.8\%)\\ \hline
\multirow{8}{*}{Jigsaw} & \multirow{4}{*}{LSTM} & country & 1000 & 22 & 26 & 106 & 123 & 224 & 254 & 32922 (72.9\%) & 123182 (88.7\%)\\
& & ethnic & 1000 & 37 & 33 & 101 & 147 & 262 & 321 & 54741 (82.0\%) & 161807 (91.7\%)\\
& & race & 1000 & 49 & 60 & 195 & 211 & 359 & 372 & 78513 (74.9\%) & 218982 (87.2\%)\\
& & religion & 1000 & 51 & 52 & 163 & 204 & 325 & 358 & 83102 (86.2\%) & 227167 (93.3\%)\\ \cline{2-12}
& \multirow{4}{*}{GRU} & country & 1000 & 22 & 34 & 84 & 95 & 232 & 250 &  39242 (74.5\%) &  98753 (88.9\%)\\
& & ethnic & 1000 & 66 & 38 & 124 & 139 & 274 & 323 & 84765 (86.3\%) & 181891 (93.3\%)\\
& & race & 1000 & 55 & 58 & 179 & 198 & 290 & 333 & 85106 (79.3\%) & 218404 (90.1\%)\\
& & religion & 1000 & 58 & 55 & 147 & 169 & 361 & 400 & 88945 (84.8\%) & 200968 (93.5\%)\\
\hline
\end{tabular}
\end{table*}

\vbox{}
\noindent \textbf{Text Data} We compare \method with random perturbation (RP) on text datasets. Table~\ref{tab:effective} presents the results where the column \emph{Protected Attr.} is the protected attribute. Column \emph{Seed} is the number of seeds in the dataset that are explored using our method. A seed is a text which contains at least one sensitive token. We limit the number of seeds to be maximum of $1,000$ to reduce the overall experiment time. If there are fewer than $1,000$ in the given set, the entire available set is used. Column \emph{Raw} shows the number of \idie s among the seeds (without any perturbation). Column \emph{RP} is the number of \idie s identified by the baseline approach where the random perturbation is applied to non-discriminatory text in the global generation, whereas \emph{\methode\_ASC} is the number of \idie s identified through gradient-guided perturbation. The results show that the \method is significantly more effective than the baseline, i.e., \textit{generating 4.40 times more \idie s on average}. In other words, gradient-guided perturbation applied to non-discriminatory text identifies much more \idie s than random searching.

Besides, we also compare \method with adversarial attacks in terms of effectiveness in generating \idie s. As shown in Table~\ref{tab:effective}, it is clear that the fairness testing method which considers two similar samples is much more effective in crafting \idie s, \textit{on average, it generates 22.11\% (18.18\%) more discriminatory text for ASC~\cite{asc} (HotFlip~\cite{hotflip}).} This is because our method \method aims to maximize the output difference between two similar samples, which effectively avoids the situation that $x$ and $x'$ are too close so that they always cross the decision boundary at the same time.

Overall, it can be observed that the number of existing \idie s or \idie s generated through perturbation on non-discriminatory text is still quite limited, i.e., no more than 211 across all datasets, all models, and all sensitive features. This suggests that the local generation is indeed necessary to further generate \idie s. For each \idi identified on the global phase\footnote{For \methode , we utilize the \idie s generated by \methode\_ASC as local generation seeds.}, we apply local generation to generate at most $1,000$ perturbed texts to identify \idie s. Column \emph{RP} and \emph{\method} show the total number of \idie s generated through random perturbation and our gradient-guided perturbation respectively. Note that we are also reporting the results of distinct texts in each entry. It is obvious that our gradient-guided perturbation generates significantly more \idie s than random perturbation, i.e., \textit{on average 2.68 times more}. Furthermore, the perturbation on a \idi by gradient-guided search has a higher success rate than the random approach. \textit{On average, 91.7\% of the distinct texts crafted by our gradient-guided perturbation are successfully identified as \idie s while the success rate of random perturbation is 82.69\%.}

We thus have the following answer to RQ1:

\begin{framed}
\noindent \emph{Answer to RQ1: \method outperforms the baseline methods. For tabular data, compared to AEQUITAS, \method searches 9.6 times input space, generates 25 times \idie s, and has more than 2 times success rate. Compared to SG, \method searches 6.6 times input space and generates 6.5 \idie s, and has a slightly higher success rate given the same time limit. Gradient provides effective guidance during both global generation and local generation. For text data, \method generates 2.68 times \idie s and has a 9.01\% more success rate than random perturbation on average.}
\end{framed}

\vbox{}
\noindent \emph{RQ2: Are \idie s generated by \method valid?}

\noindent Recall that we aim to generate valid \idie s, especially we need to preserve the syntax and semantics for the text. To answer this question, we measure how close the generated \idie s are to the original ones using multiple metrics that are commonly applied in NLP research. Noted that these metrics also can be utilized to filter the similar text by giving a threshold~\cite{filter}.
\begin{itemize}
    \item \emph{$L_p$ Norm Distance} is widely used to assess the dissimilarity of two samples. Specifically, $L_0$ distance counts the number of tokens modified and $L_2$ distance quantifies the Euclidean distance in the word vector space. A smaller distance indicates higher similarity.
    \item \emph{Jaccard Similarity Coefficient} is another popular statistical indicator to measure similarity. Given two sets containing all tokens in the two texts respectively, it is defined as the division of the size of the intersection over the size of the union. The larger is the coefficient, the better.
    \item \emph{BLEU Score}~\cite{bleu} is widely used in neural machine translation to measure the differences between a candidate and reference sentences~\cite{nmt}. It first calculates the $N$-grams ($N=1,2,3,4$) model of the two texts respectively and then counts the number of matches. In our experiment, it is used to evaluate the quality of the generated IDS by regarding the original text as a reference. The score ranges from 0 to 1 and a higher score is preferred.
    \item \emph{Semantic Similarity} converts words to vectors using an embedder and uses their cosine similarity for the evaluation. In our experiment, we utilized the model in~\cite{sentence_encoder} to convert the original text and generated IDS to two 512-dimensional vectors and then compute the cosine of their angles. The higher the score is, the better.
\end{itemize}

\begin{figure*}[t]
\subfigure[$L_0$ Norm Distance]{
\includegraphics[width=0.18\textwidth]{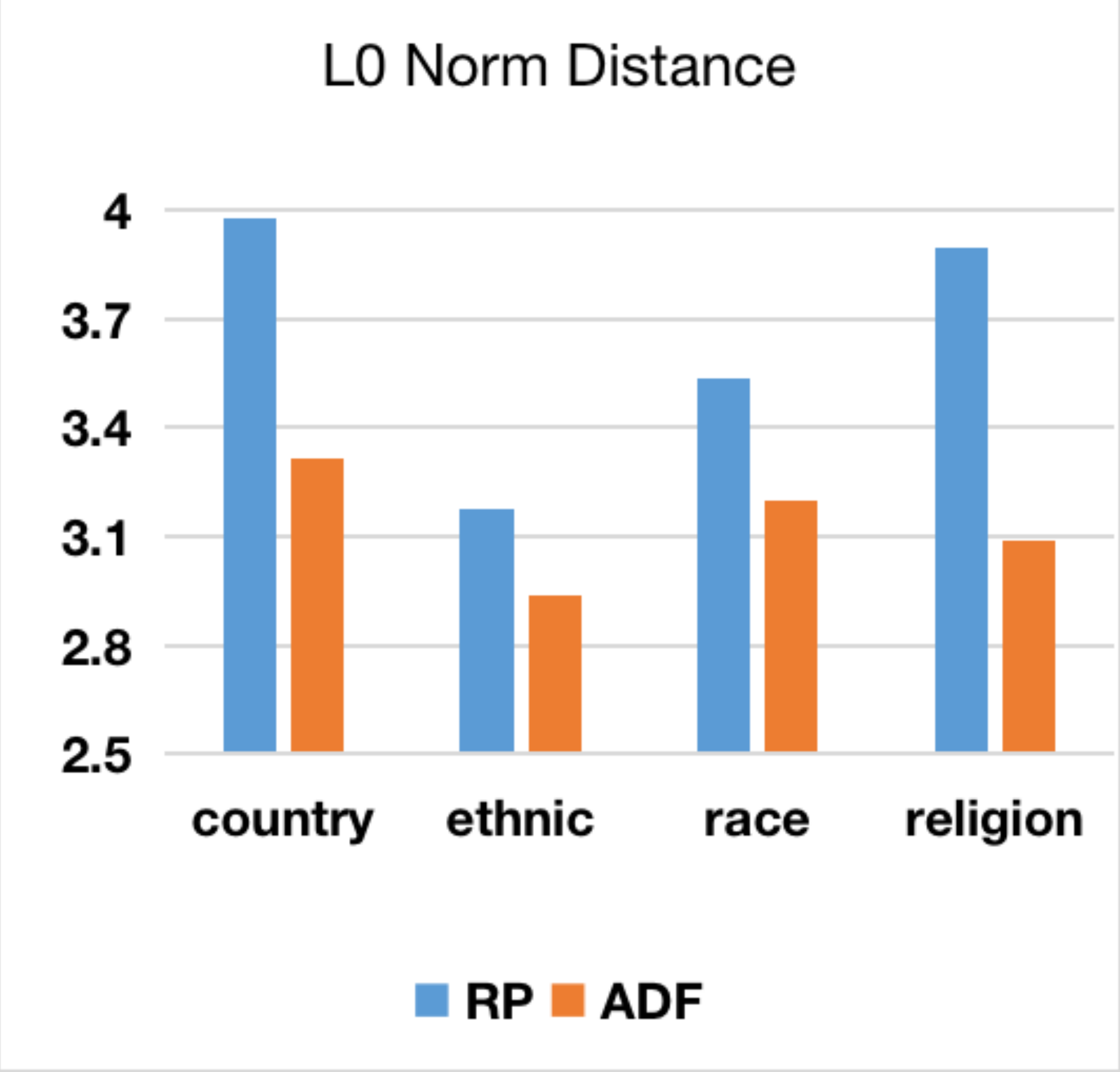}
\label{fig:l0d}
}
\subfigure[$L_2$ Norm Distance]{
\includegraphics[width=0.18\textwidth]{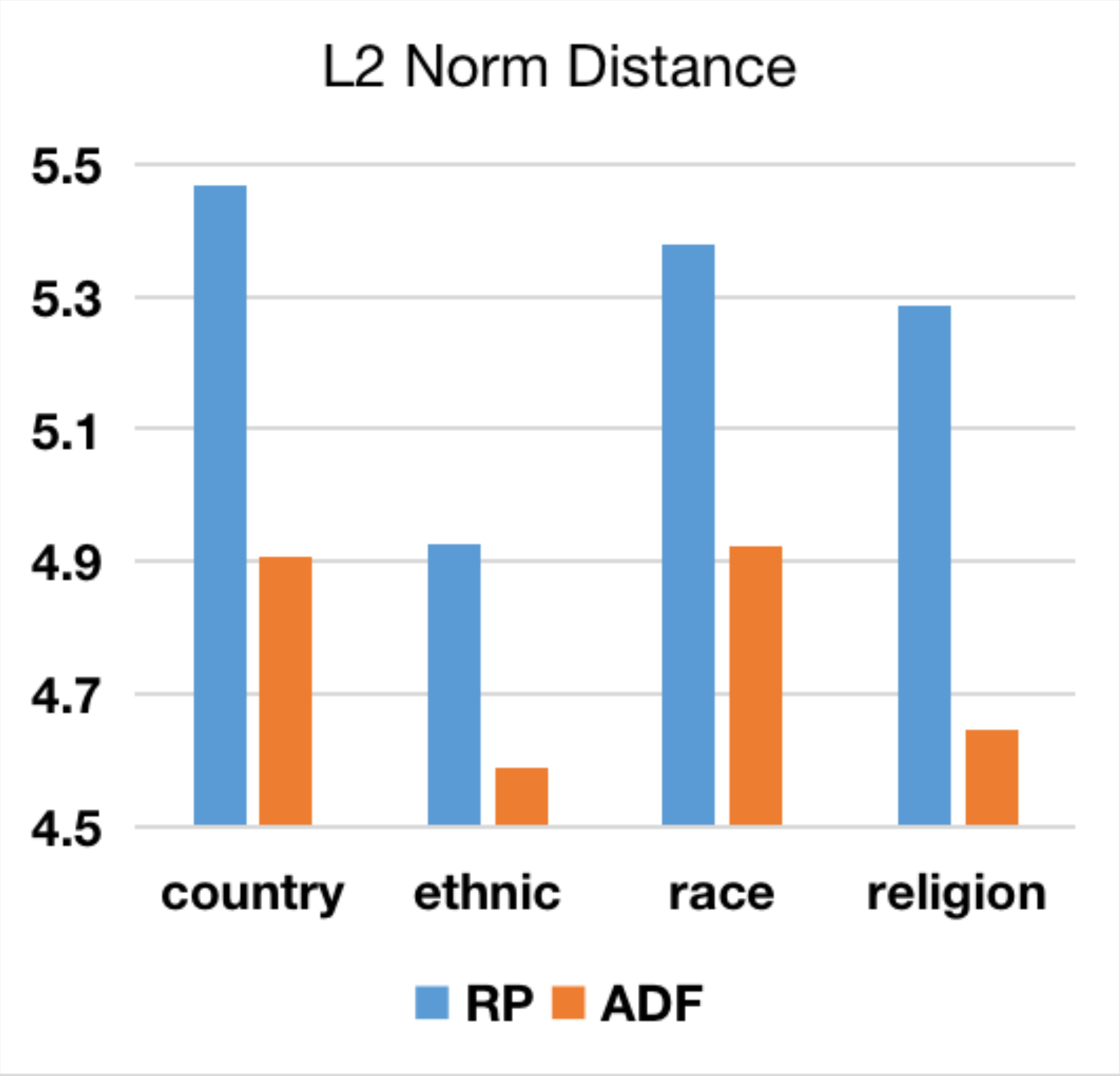}
\label{fig:l2d}
}
\subfigure[Jaccard Similarity Coefficient]{
\includegraphics[width=0.18\textwidth]{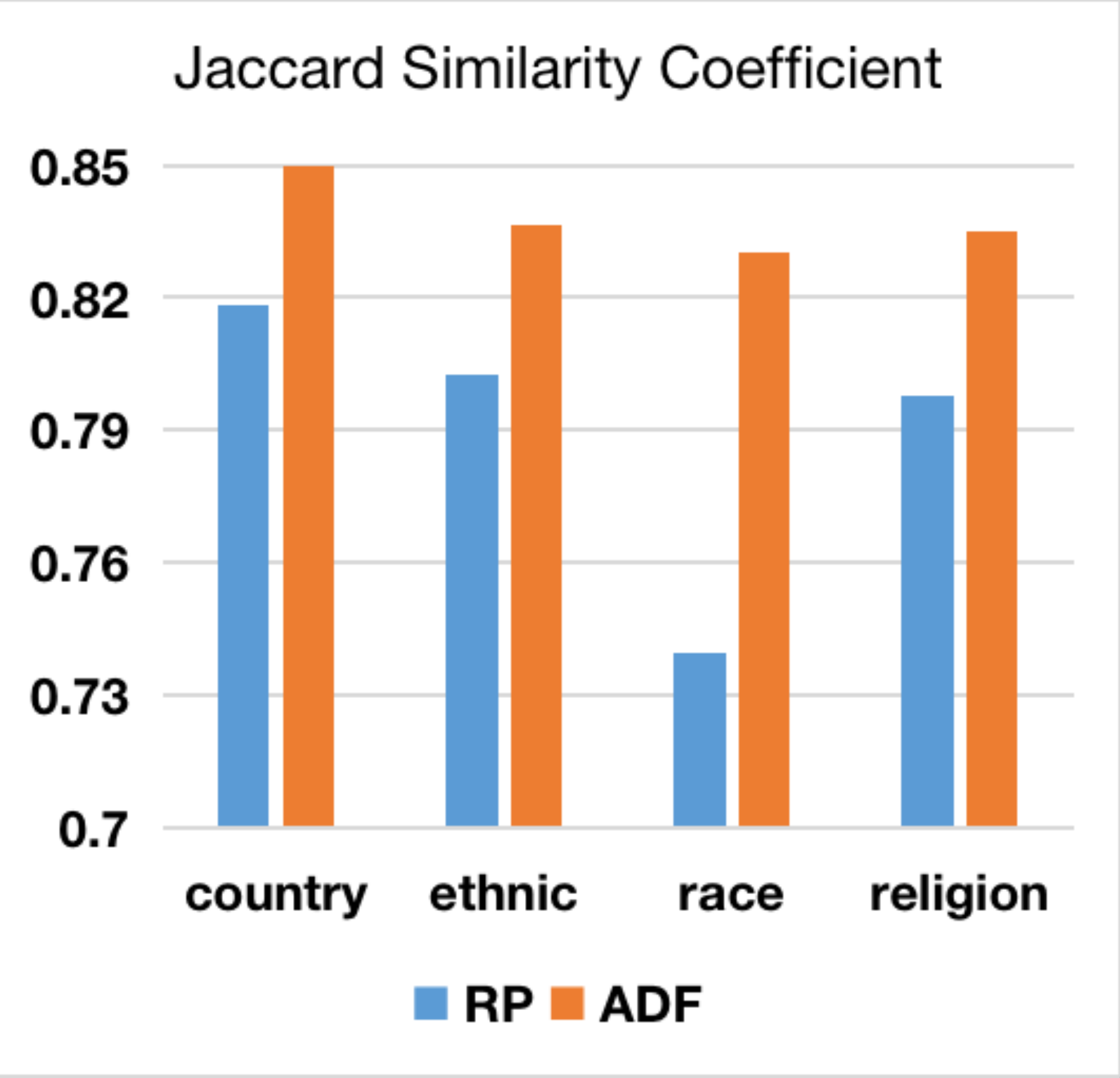}
\label{fig:jsc}
}
\subfigure[BLEU Score]{
\includegraphics[width=0.18\textwidth]{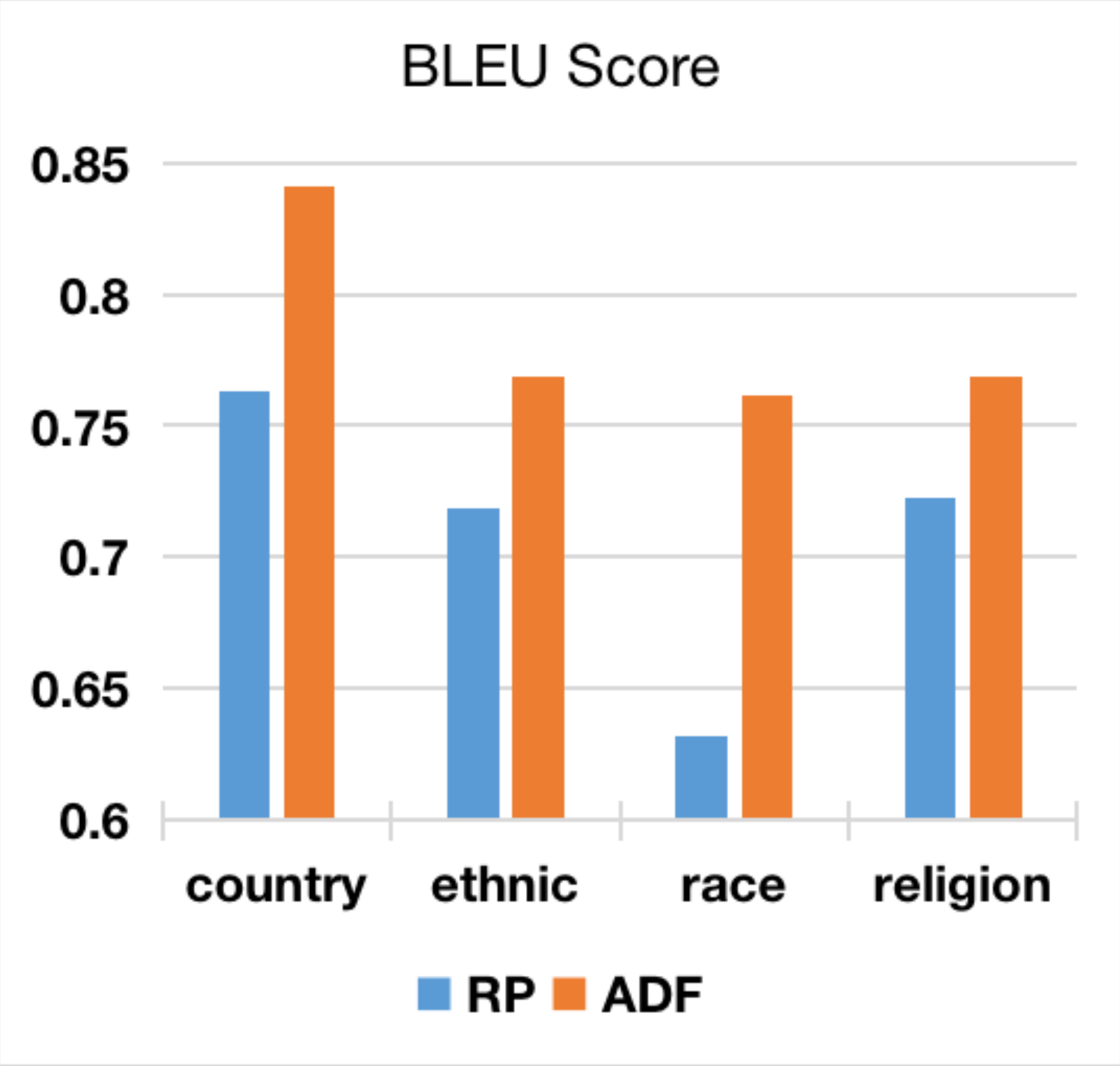}
\label{fig:bleu}
}
\subfigure[Semantic Similarity]{
\includegraphics[width=0.18\textwidth]{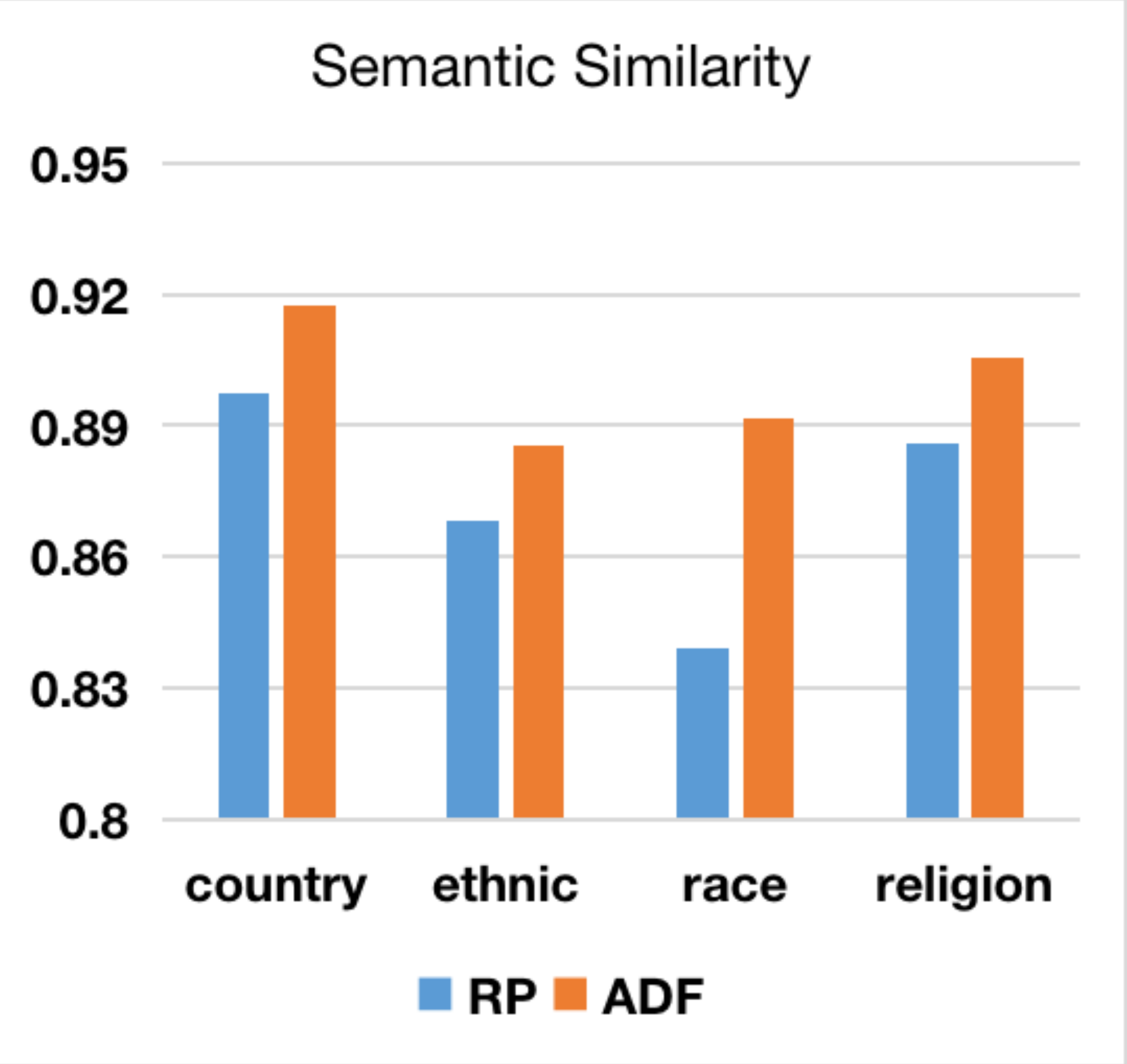}
\label{fig:ss}
}
\caption{Validity analysis.}
\label{fig:ma}
\end{figure*}

For a fair comparison on text datasets with the baseline, we take the same number of \idie s identified on the global generation phase and generate \idie s by local generation. Afterward, we compute and compare the above-mentioned metrics for both approaches. Figure~\ref{fig:ma} shows the average metric values computed based on the LSTM model trained on the Wiki Comment dataset. Since the results of other models and other datasets are similar, they are omitted and the readers are referred to~\cite{appendix} for details. 

First, it can be observed that a consistent good score is achieved by \idie s generated by our approach across all five metrics. For instance, the average $L_0$ norm distance is 3.13, which intuitively means that only 3.13 tokens on averages are modified. Second, it is evident that our gradient-guided perturbation performs better than random perturbation on all metrics, i.e., the \idie s generated by our approach are more similar with the seed text than those generated through random perturbation. \textit{On average, \idie s generated by our approach modify 13.55\% less tokens (see Figure~\ref{fig:l0d}), have 9.39\% shorter distance in word embedding space (see Figure~\ref{fig:l2d}), and are 7.25\%, 3.17\%, 11.04\% more similar to the original text in terms of Jaccard Similarity Coefficient (see Figure~\ref{fig:jsc}), Semantic Similarity (see Figure~\ref{fig:ss}) and BLEU Score (see Figure~\ref{fig:bleu}) respectively.}

\begin{table}[t]
\centering
\caption{Validity analysis w.r.t. different global generation strategies.}
\label{tab:va_ft}
\begin{tabular}{|c|c|c|c|c|}
\hline
Metric & ASC & \methode\_ASC & HF & \methode\_HF\\
\hline
L\_0 & 2.6296 & 2.1851 & 2.6364 & 2.6136\\ \hline
L\_2 & 5.5460 & 5.2825 & 5.8291 & 5.5203\\ \hline
JSC & 0.8514 & 0.8770 & 0.8624 & 0.8724\\ \hline
BLEU & 0.8186 & 0.8577 & 0.8482 & 0.8634\\ \hline
SS & 0.8914 & 0.9135 & 0.8862 & 0.8967\\ \hline
\end{tabular}
\end{table}

In Table~\ref{tab:va_ft}, we conduct an experiment to evaluate the quality of \idie s generated by different global generation strategies. Similarly, only the result of the LSTM model trained on the Wiki Comment dataset (protected attribute is \emph{race}) is shown here, and the remaining results are in~\cite{appendix}. The results show that for ASC (HotFlip), comparing with the adversarial attack version, \textit{\method generates \idie s with 0.44 (0.02) fewer perturbed tokens, 0.26 (0.31) shorter Euclidean distance, and 0.03 (0.01), 0.02 (0.01), 0.04 (0.02) more similar to the original text in terms of Jaccard Similarity Coefficient, Semantic Similarity and BLEU Score respectively.}

We thus have the following answer to RQ2:

\begin{framed}
\noindent \emph{Answer to RQ2: The \idie s generated through \method are similar to the original one.}
\end{framed}

\vbox{}
\noindent \emph{RQ3: How efficient is our algorithm in finding \idie s?}

\noindent Besides effectiveness, efficiency is also important. We thus conduct an experiment to compare the efficiency of \method with baseline.  

\begin{table}[t]
\centering
\caption{Time (seconds) taken to generate 1,000 \idie s on tabular datasets.}
\label{tab:efficiency_tabular}
 \resizebox{.4\textwidth}{!}{
\begin{tabular}{|c|c|c|c|c|}
\hline
Dataset & Prot. Attr. & AEQUITAS & SG & \method \\
\hline
census & age & 172.64 & 720.49 & 59.15\\
census & race & 128.75 & 506.33 & 65.95\\
census & gender & 158.37 & 2128.42 & 78.68\\
bank & age & 191.16 & 521.79 & 106.93\\
credit & age & 176.31 & 321.63 & 64.92\\
credit & gender & 156.22 & 476.52 & 102.90\\
\hline
\end{tabular}
 }
\end{table}

\begin{table}[t]
\centering
\caption{Time (seconds) taken to generate 5,000 \idie s on text datasets.}
\label{tab:efficiency_text}
\begin{tabular}{|c|c|c|c|c|}
\hline
Dataset & Model & Prot. Attr. & RP & \methode\_ASC \\
\hline
\multirow{8}{*}{Wiki} & \multirow{4}{*}{LSTM} & country & 2894.8 & 2371.5\\
& & ethnic & 563.9 & 837.2 \\
& & race & 659.0 & 1587.4 \\
& & religion & 542.0 & 746.8 \\ \cline{2-5}
& \multirow{4}{*}{GRU} & country & 2037.8 & 746.8 \\
& & ethnic & 1307.4 & 1142.6 \\
& & race & 451.7 & 638.3 \\
& & religion & 510.9 & 551.1 \\ \hline
\multirow{8}{*}{Jigsaw} & \multirow{4}{*}{LSTM} & country & 2069.5 & 1848.1 \\
& & ethnic & 738.3 & 1136.9 \\
& & race & 1296.3 & 2677.5 \\
& & religion & 444.2 & 685.6 \\ \cline{2-5}
& \multirow{4}{*}{GRU} & country & 1572.1 & 910.6 \\
& & ethnic & 1265.5 & 2267.6 \\
& & race & 472.6 & 583.1 \\
& & religion & 919.8 & 438.5 \\
\hline
\end{tabular}
\end{table}

Table~\ref{tab:efficiency_tabular} presents how much time each method takes to generate 1,000 \idie s on tabular datasets. Note that for all methods we measure the total time, especially for SG, it includes the time for generating the explanation model and constraint solving. It is evident that \method has the best performance. \textit{On average, it takes only 48.97\% and 14.53\% of the time required by AEQUITAS and SG respectively.} Combined with the results shown in Table~\ref{tab:comparison_aequitas}, it implies that AEQUITAS and ADF have similar efficiency in generating samples (and ADF has a much higher success rate in finding \idie s). Considering that AEQUITAS performs random sampling whereas \method needs to calculate the gradient, it suggests that the overhead of calculating gradient in ADF is negligible. SG takes significantly more time to generate samples based on a seed sample. Its efficiency is thus much worse, as expected. 

For the text datasets, we generate 5,000 \idie s by both the baseline approach and our approach (\methode\_ASC) and record the time taken in Table~\ref{tab:efficiency_text}. Overall, \method takes an average of 1234.2 seconds across all datasets and models, which we believe are reasonably efficient, compared to the cost of data collection and model training for such tasks. Comparing our approach with the baseline, \textit{RP takes on average $2.9\%$ less time than our approach.} The additional time overhead is mainly due to the extra step of calculating the gradients on both global and local generation stages. Due to the structure of RNN, the time complexity of back-propagation by chain-rule is $\mathbf{O}(N)$, where $N$ is the length of the text. Further experiment shows that this step is the most time consuming, e.g., in the case of the LTSM model trained on the Wiki Comments, each gradient computation takes 0.454 (0.184) seconds, while the remaining steps only take 0.060 (0.041) seconds for perturbation on global (local) phase. Besides, we also compare the efficiency of \methode\_ASC and \methode\_HF, they take an average of $0.83$ and $3.95$ seconds each iteration respectively. This is because HotFlip traverses all possibilities of perturbed and substituting tokens in each iteration, which makes its time complexity is $\mathbf{O}(m*k)$, where $k$ is the number of similar tokens for selecting. However, the time complexity of ASC is $\mathbf{O}(k)$, since it greedily chooses the most important token and then only traverses the top-$k$ similar words.

We thus have the following answer to RQ3:
\begin{framed}
\noindent \emph{Answer to RQ3: \method is reasonably efficient in generating \idie s. There is a slight time overhead in applying gradient-guided perturbation, which is more noticeable for RNN.}
\end{framed}

\vbox{}
\noindent \emph{RQ4: How useful are the identified \idie s for improving the fairness of the DNN?}

\begin{table}[t]
\centering
\caption{Fairness improvement on tabular datasets.}
\label{tab:improvement_tabular}
\begin{tabular}{|c|c|c|c|c|c|}
\hline
\multirow{2}{*}{Dataset} & \multirow{2}{*}{Prot. Attr.} & \multirow{2}{*}{Before (\%)} & \multicolumn{3}{c|}{After (\%)} \\
\cline{4-6} & & & AEQUITAS & SG & \method \\
\hline
census & age & 10.88 & 4.03 & 2.41 & 2.26\\
census & race & 9.75 & 7.05 & 6.89 & 6.15\\
census & gender & 3.14 & 2.33 & 1.90 & 1.65\\
bank & age & 4.60 & 1.68 & 2.04 & 1.19\\
credit & age & 27.93 & 13.91 & 13.19 & 12.05\\
credit & gender & 7.68 & 4.58 & 4.66 & 3.93\\
\hline
\end{tabular}
\end{table}

\noindent To further show the usefulness of our generated \idie s, we evaluate whether we can improve the fairness of the DL model by retraining it with data augmented with the generated \idie s. We remark that \aeq also uses retraining to improve the fairness of the original models and SG does not have such discussions. We need a systematic way of evaluating the fairness of a given model. For this, we adopt the method proposed and used by \aeq ~\cite{aequitas}. The idea is to randomly sample a large set of samples and evaluate the model fairness by the percentage of \idie s in the set. However, as it is nearly impossible to obtain meaningful text by random sampling, we evaluate the improvement by an independently generated set of \idie s, which is a common practice~\cite{deeptest,surprise}. That is, we apply our approach to generate an independent set of IDSs based on the testing set and evaluate how many of them are IDSs with respect to the retrained model. Note that since we randomly select 5\% of generated \idie s for data augment and retraining, we repeated the procedure 5 times and present the average improvement.

The results on tabular datasets are shown in Table~\ref{tab:improvement_tabular}, where columns \emph{Before} and \emph{After} are the estimated fairness of the model before and after retraining using the generated \idie s. The smaller the number is, the more fair the model is. It can be observed that retraining with the \idie s can significantly improve the fairness, and \textit{\method achieves better fairness improvement (with more identified \idie s), i.e., 57.2\% on average, versus existing approaches, i.e., 45.1\% for AEQUITAS and 49.1\% for SG.}

\begin{table}[t]
\centering
\caption{Fairness improvement on text datasets.}
\label{tab:improvement_text}
\begin{tabular}{|c|c|c|c|c|}
\hline
Dataset & Model & Prot. Attr. & Augmented (\%) & Certified (\%)\\
\hline
\multirow{8}{*}{Wiki} & \multirow{4}{*}{LSTM} & country & 72.4 & 54.9\\
& & ethnic & 70.2 & 64.1\\
& & race & 64.2 & 61.8 \\
& & religion & 54.2 & 63.4\\ \cline{2-5}
& \multirow{4}{*}{GRU} & country & 48.6 & 66.3\\
& & ethnic & 66.2 & 71.0\\
& & race & 58.1 & 65.8\\
& & religion & 51.0 & 59.8\\ \hline
\multirow{8}{*}{Jigsaw} & \multirow{4}{*}{LSTM} & country & 71.6 & 51.5\\
& & ethnic & 70.4 & 55.6\\
& & race & 58.5 & 60.4\\
& & religion & 63.8 & 55.5\\ \cline{2-5}
& \multirow{4}{*}{GRU} & country & 64.1 & 57.9\\
& & ethnic & 56.6 & 61.4\\
& & race & 53.5 & 66.3\\
& & religion & 40.4 & 59.8\\
\hline
\end{tabular}
\end{table}

\begin{table}[t]
\centering
\caption{Fairness improvement w.r.t. different methods. The improvement of \methode\_ASC is showed in Table~\ref{tab:improvement_text}.}
\label{tab:improvement_ft}
\begin{tabular}{|c|c|c|c|c|c|}
\hline
Dataset & Model & Prot. Attr. & ASC & HF & \methode\_HF\\
\hline
Wiki & LSTM & race & 63.4 & 66.3 & 67.5 \\ \hline
Jigsaw & GRU & ethnic & 57.1 & 61.7 & 62.0 \\
\hline
\end{tabular}
\end{table}

Table~\ref{tab:improvement_text} presents the results on text datasets. Note that the augmented data is crafted by \methode\_ASC based on the training dataset, which has no overlap with the testing dataset. The number of independently generated \idie s is 155,690 on average for each row. Column \emph{Augmented} shows the percentage of those independently generated \idie s that are no longer \idie s given the retrained model. That is, the bigger the number, the less discriminative the retrained model. As observed from Table~\ref{tab:improvement_text}, \textit{retraining reduces the number of \idie s by an average of 60.2\% and up to 72.4\%}. Since individual fairness can be regarded as a specific form of robustness according to its definition, we also compare the effectiveness of augmented training with certified training which acquires the tractable upper bound for the worst-case perturbation and then provides a certificate of robustness ~\cite{certified1,certified3,certified2}. Since ~\cite{certified3,certified2} propagate the upper bound with interval arithmetic proposed by ~\cite{certified4}, which is only sufficient for FNNs and CNNs, here we adopt ~\cite{certified1} to present a comprehensive analysis. Column \emph{Certified} listed the fairness improvement for the certified trained model. \textit{On average, the performances of augmented training and certified training achieve similar improvement on fairness, 60.2\% versus 61.0\%.} Although certified training is a one-time effort for all the potential sensitive tokens, fairness testing not only improves the fairness of the model but also quantifies the bias of the original model. In addition, we conduct a supplementary experiment to show the fairness improvement with regard to different sample generating methods in Table~\ref{tab:improvement_ft}, and the results of \methode\_ASC are presented in Table~\ref{tab:improvement_text}. Noted that the testing data is all generated using method \methode\_ASC. It reveals that the greater the number and diversity of \idie s, the more helpful to improve the fairness of the model.

We thus have the following answer to RQ4:
\begin{framed}
\noindent \emph{Answer to RQ4: The \idie s generated by \method are useful to improve the fairness of the DL models through retraining, with an average improvement of 57.2\% on tabular datasets and 60.2\% on text datasets.}
\end{framed}

%% file: 5_Discussion.tex
\subsection{Threats to Validity}
\label{sec:threat}

\noindent\textbf{Limited datasets}
In the experiment, We evaluated \method with \datasets datasets, including \textdata text datasets for toxic classification tasks. Compared with other text classification tasks, toxic classifiers are more prone to discrimination issues, e.g., racism, sexism, and most of the relevant data can be easily obtained from social media. Although they are the most common public benchmarks used in the fairness testing literature, we cannot conclude the effectiveness and efficiency of other datasets. As our approach is independent of the datasets and has been made available online, it can be used on other datasets with minor adjustments for data adaptation.

\vbox{}
\noindent\textbf{Limited models}
We only tested three deep learning models in the experiment. Especially for the tabular data, since they are relatively simple (i.e., with a maximum of 20 features), we only used the basic fully connected DNN. However, the key idea of \method is generic which can be easily implemented for more complex deep learning models like convolutional neural networks (CNNs), as our approach only requires a way of computing the gradients.

\vbox{}
\noindent\textbf{White-box setting}
\method is designed as a white-box approach, i.e., the fairness testing part requires the full knowledge of the target deep learning models. It is widely accepted that deep learning model testing could have the full knowledge of the target model. In the future, we will explore how to extend it to a black-box setting, e.g., selecting the token for perturbation not based on gradients but certain importance scores (i.e., the confidence of prediction).

\vbox{}
\noindent\textbf{Step-size parameters}
The step-size parameters of \method for tabular data depend on the training dataset. For datasets with only categorized attributes, it is easy to set it to be 1. For other datasets, further research may be necessary to identify an effective step size. If the step size is too big, it may miss some \idie s during its perturbation, especially for the local generation. If the step size is too small, it is hard to generate \idie s in the global generation.

\vbox{}
\noindent\textbf{Complex context}
Text data is much more complex than tabular data and it is certain that our approach does not cover all kinds of discrimination. First, one word could have different meanings according to their contexts, for example, ``black'' and ``white'' may refer to either color or race. We thus are unable to filter them out without manual labeling. Second, there may be many forms of specific discrimination. For instance, sexism may not only be associated with tokens such as ``male/female'', but also ``he/she'', and ``actor/actress''. It is thus nearly impossible to list them all.

\subsection{Discussion}
\label{sec:discussion}

In this work, we aim to acquire a large number of diverse discriminatory samples, since the diversity and quantity of \idie s would help us to 1) figure out whether there is discrimination in different subspaces of the model, 2) mitigate the discrimination in multiple parts of the model at one time through retraining. Clustering the original data in \method can be regarded as a coarse-grained equivalence class partition method since perturbation only slightly shifts the distribution. However, this is not enough, we hope that there will be more fine-grained methods for dividing equivalence classes, which could help us better understand the causes of discrimination and improve model fairness.

Another point worth discussing is the tolerance of fairness. Due to the sensitivity of fairness in society, fairness is strictly required in many cases. For example, many USA states prohibit the use of face recognition in public places recently, and discrimination is one of the most important reasons. Although it is challenging to achieve complete fairness for DL models, our goal is to improve the fairness of DL models, rather than giving up DL models. First, DL does bring convenience to our daily life. Second, manual decision-making is not completely fair either. The real problem is that discrimination in historical data may be introduced or even magnified in the training process of the model. Thus, we need to locate the data on which discrimination occurs and figure out how to reduce the discrimination learned by the model as much as possible, and so that its fairness can reach the same level as or even higher than that of manual decision-making.

%% file: 6_Related_Work.tex
\section{Related Work}
\label{sec:relatedwork}

\noindent\textbf{Fairness}
Many works exist on the fairness issues of AI in general. In~\cite{why}, Chen \emph{et al.} attributed the unfairness of the trained model to the data collection, then decomposed the discrimination of data into bias, variance, and noise, and last proposed some strategies to estimate and reduce them. In~\cite{fat}, Albarghouthi \emph{et al.} proposed fairness-aware programming, which monitors whether programs violate custom fairness at runtime by concentration inequalities~\cite{ci}. In~\cite{verification}, Bastani \emph{et al.} first formalized three fairness specifications demographic parity, equal opportunity, and path-specific causal fairness, and then utilized adaptive concentration inequalities to obtain a probabilistic guarantee of the model with respect to the given specification. In~\cite{science}, Thomas \emph{et al.} proposed a general and flexible framework by encoding fairness constraints (upper bound) into the objective function. When the amount of data is enough, Hoeffding’s inequality~\cite{inequality} is used to obtain the bounds, otherwise, Student's t test is used. In addition to these general methods, some works are focused on the fairness of text. In~\cite{discuss_1,discuss_2}, the impacts of using NLP models and potential bias caused in real-world applications are discussed. \cite{embedding} points out that the bias may exist in word embeddings. It also proposes a mechanism on gender-neutral words to quantify the degree of gender bias by projection and reduces the bias by removing the gender associations. \cite{unintended_bias} shows that the discrimination is correlated with comment length and the distribution differences between the sensitive values in the toxic data and the whole dataset. In~\cite{counterfactual}, Garg \emph{et al.} improve the model's fairness through a robust training method called CLP which introduces the prediction difference of the original input and its counterpart (which only differs in certain sensitive features) as penalties into the training loss.

\vbox{}
\noindent\textbf{Fairness testing}
This work is closely related to fairness testing of machine learning models. Galhotra \emph{et al.} proposed \tm ~\cite{themis} which firstly defines software fairness testing, then introduces fairness scores as measurement metrics and lastly designs a causality-based algorithm utilizing the random test input generation technique to evaluate the model fairness, i.e., the frequency of \idie s' occurrence of software. However, \tm is inefficient in general since it relies on random sampling without guidance on the generation. Udeshi \emph{et al.} proposed \aeq ~\cite{aequitas} which inherits and improves \tm and focuses on the \idi generation. \aeq is a systematic generating algorithm. It first explores the input domain randomly to discover \idie s in the global search phase. During the local generation, \aeq searches the neighbors of \idie s identified in the global phase, by perturbing them. For the local generation, \aeq designs three different strategies, i.e., random, semi-directed, and fully-directed, to update the probability which is used to guide the selection of attributes to perturb. Based on their evaluation, fully-directed has the best effectiveness and efficiency. Besides searching the \idie s, \aeq also design an automated iterative retraining method to obtain a more fair model. Later, Agarwal \emph{et al.} proposed Symbolic Generation (SG) ~\cite{sg} which integrates symbolic execution and local model explanation techniques to craft \idie s. SG relies on the local explanation of a given input which constructs a decision tree utilizing the samples generated randomly by the Local Interpretable Model-agnostic Explanation (LIME) ~\cite{lime}. The path of the tree determines all the important attributes leading to the prediction. The algorithm also contains a global generation phase and a local generation phase. A detailed comparison between ADF and the above approaches is presented in Section~\ref{qualitative}.

\vbox{}
\noindent\textbf{Gradient-based attacks}
This work is also related to research on gradient-based adversarial attacks. A variety of works have been proposed to explore the vulnerability of the DL model by crafting adversarial samples, and gradient-based adversarial attack is one kind of the most effective methods. Goodfellow \emph{et al.} proposed the first attacking algorithm Fast Gradient Sign Method (FGSM)~\cite{fgsm} to generate adversarial samples by perturbing the original input with the linearization of the loss function used in the training process. FGSM is fast by only attacking once according to the gradient. Later, several other attack methods are proposed to extend FGSM. For instance, instead of attacking only once, Basic Iterative Method (BIM)~\cite{bim} employs perturbations based on gradients multiple times (often with smaller step sizes) and applies a function that performs per-attribute clipping to make sure the sample after each iteration is located in the neighborhood of the original sample. In ~\cite{asc}, Papernot \emph{et al.} first formalize the adversarial sequence optimization problem in the context of sequential data and develops ASC to obtain malicious texts by iteratively replace a token with the one such that the sign of the difference between them is closest to the sign of gradient. However, ASC may cause syntactic or semantic errors since it searches in the whole corpus. In ~\cite{textbugger}, a general attack framework called TEXTBUGGER is proposed to generate adversarial texts. It works as follows: 1) find the important words by a) computing the Jacobian matrix under white-box setting and b) scoring with the confidence drop before and after removal of a word under black-box setting; 2) generate adversarial texts by inserting a space, deleting the middle letter, swapping two adjacent characters, replacing a letter with a similar one, and taking the top$-k$ nearest words in embedding space as a substitute. In ~\cite{hotflip}, HotFlip is proposed to generate adversarial examples with character/word substitution, which uses beam search to apply perturbation with the highest loss estimated by gradient with respect to the one-hot input. Besides, Pei \emph{et al.} ~\cite{deepxplore} designed an algorithm for maximizing the coverage of neurons as well as model outputs of multiple DNNs, and solve the optimization function using the gradient.

%% file: 7_Conclusion.tex
\section{Conclusion}
\label{sec:conclusion}
In this work, we propose a lightweight algorithm \method to efficiently generate \idie s for deep learning models through adversarial sampling. Our algorithm combines a global phase and a local phase to systematically search the input space for \idie s with the guidance of gradient. In the global generation, \method first locates the \idie s near the decision boundary by iteratively perturbing towards the decision boundary. In the local generation, \method again samples according to the gradient to search the neighborhood of a found \idie. The experiment on \benchmarks benchmarks of \datasets datasets shows that \method is able to effectively generate diverse valid \idie s within a reasonable time. 

As for future work, we would like to propose additional criteria to measure the generated \idie s, and then cluster them into different equivalence classes to improve the effectiveness of augmented retraining or fine-tuning. Besides, we also want to uncover the root cause of discrimination in the model, e.g., mapping it to the training data.